\documentclass{vgtc}                          % final (conference style)
\ifpdf%                                % if we use pdflatex
  \pdfoutput=1\relax                   % create PDFs from pdfLaTeX
  \usepackage{graphicx}                % allow us to embed graphics files
  \DeclareGraphicsExtensions{.pdf,.png,.jpg,.jpeg} % for pdflatex we expect .pdf, .png, or .jpg files
\else%                                 % else we use pure latex
  \usepackage{graphicx}                % allow us to embed graphics files
  \DeclareGraphicsExtensions{.eps}     % for pure latex we expect eps files
\fi%

\graphicspath{{figures/}{pictures/}{images/}{./}} % where to search for the images

\usepackage{microtype}                 % use micro-typography (slightly more compact, better to read)
\PassOptionsToPackage{warn}{textcomp}  % to address font issues with \textrightarrow
\usepackage{textcomp}                  % use better special symbols
\usepackage{mathptmx}                  % use matching math font
\usepackage{times}                     % we use Times as the main font
\usepackage{cite}                      % needed to automatically sort the references
\usepackage{tabu}                      % only used for the table example
\usepackage{booktabs}                  % only used for the table example
\onlineid{0}

\vgtccategory{Research}

\vgtcinsertpkg

\title{A Weakly-Supervised Attention-based Visualization Tool for Assessing Political Affiliation}

\author{Srijith Rajamohan\thanks{e-mail: srijithr@vt.edu}\\ %
        \scriptsize Virginia Tech %
\and Alana Romanella\thanks{e-mail: aromanel@vt.edu}\\ %
     \scriptsize Virginia Tech %
\and Amit Ramesh\thanks{e-mail: amit@vt.edu}\\ %
     \parbox{1.4in}{\scriptsize \centering Virginia Tech}}

\abstract{In this work, we seek to finetune a weakly-supervised expert-guided Deep Neural Network (DNN) for the purpose of determining political affiliations. In this context, stance detection is used for determining political affiliation which is framed in the form of relative proximities between entities in a low-dimensional space. An attention-based mechanism is used to provide model interpretability. A Deep Neural Network for Natural Language Understanding (NLU) using static and contextual embeddings is trained and evaluated. Various techniques to visualize the projections generated from the network are evaluated for visualization efficiency. An overview of the pipeline from data ingestion, processing and generation of visualization is given here. A web-based framework created to faciliate this interaction and exploration is presented here. Preliminary results of this study are summarized and future work is outlined.%
} % end of abstract

\CCScatlist{
  \CCScatTwelve{Human-centered computing}{Visu\-al\-iza\-tion}{Augmented Intelligence}{};
  \CCScatTwelve{Human-centered computing}{Deep Learning}{Natural Language Processing}{}
}

\begin{document}

%% The ``\maketitle'' command must be the first command after the
%% ``\begin{document}'' command. It prepares and prints the title block.

%% the only exception to this rule is the \firstsection command
%\firstsection{Introduction}

\maketitle

\section{Introduction} %for journal use above \firstsection{..} instead

Research has been done using Deep Neural Networks (DNN) for sentiment analysis of various corpora, ranging in length from a sentence to an essay. DNNs \cite{nn2,nn3,nn4,nn5,nn6,nn7} have had success with estimating discrete levels of sentiment in short texts while variational statistics-based methods such as Latent Dirichlet Allocation (LDA) \cite{lda} work better for longer texts \cite{ldacomparison}. Most works involving political affiliation or stance detection employ a strategy of determining a discrete score for a specific set of target topics \cite{nn2,nn4}. In this work, we propose a different method where affiliation is measured as the proximity of a projected entity in a low-dimensional space to other entities. This approach has two distinct benefits: a list of topics does not have to be maintained and relative affiliations can be obtained without the need for explicit quantification of individual topic-based stances.

Kosinski et. al. \cite{kosinski} covered the extraction of stance from an extensive list of attributes obtained from Facebook. They performed this using attributes that most people assume to be private. However, their approach utilized a dichotomous political affiliation, i.e. binning users as Democrat or a Republican which we try to avoid here in our work. This paper, however, demonstrates how information about even a single attribute results in non-negligible accuracy for their predictions. Following up on this line of work, we seek to determine if information extracted from just a Twitter User's Description (TUD) using techniques based on Deep Learning and NLP is sufficient to ascertain their affiliation.

Other work involved that by Makazhanov \cite{makazhanov} which used tweets from a user and their interaction with others which was then classified using Naive Bayes with their class probabilities. Along with Naive Bayes, they also used logistic regression and decision trees. 
Mohammad et al. \cite{mohammad} created an interactive visualization, however this summarizes the makeup of the data as opposed to exploring the relationship between documents as is done in our work presented here. It must be pointed out here that Mohammad et al. also mentions distant supervision which is similar to the weak supervision that is performed here. However, Mohammad et al. uses hashtags for the distant supervision while we use techniques based on Natural Language Processing or more specifically Natural Language Understanding to perform the weak supervision.
Conover et al. \cite{conover} used Latent Semantic Analysis to obtain two-dimensional projections from a user-hashtag feature matrix. In this work, we use a DNN to train a weakly-supervised model and computes a lower-dimensional projection from the penultimate layer in the DNN using a variety of dimensionality reduction techniques.

This work focuses on the issue of detecting stance for the purpose of determining political affiliation, in particular by only looking at the contents of a Twitter User's Description (TUD) as opposed to looking at the contents of the tweet text. We accomplish this by training a DNN with a noisy-labeled corpus for binary classification. After training, a test set is fed to the trained network which results in a lower-dimensional projection extracted from the penultimate layer of the network. The projections can be visualized in two or three dimensions using one of the many visualization techniques although Multidimensional Scaling (MDS) \cite{mds} and t-SNE \cite{tsne} are utilized here. Apart from these two techniques, Principal Component Analysis (PCA) and Isomap \cite{isomap} were also evaluated and can be swapped in for either of the above projections. The suitability of all these four techniques with respect to our goals will be assessed. Two-dimensional projections are used here since it is generally easier to determine distances between projected entities in a two-dimensional plane. The end-user interacts with these projections and provides input on the generated labels which is used to further train and finetune the network. 

\begin{figure}[h]
    \centering
    \includegraphics[width=\linewidth]{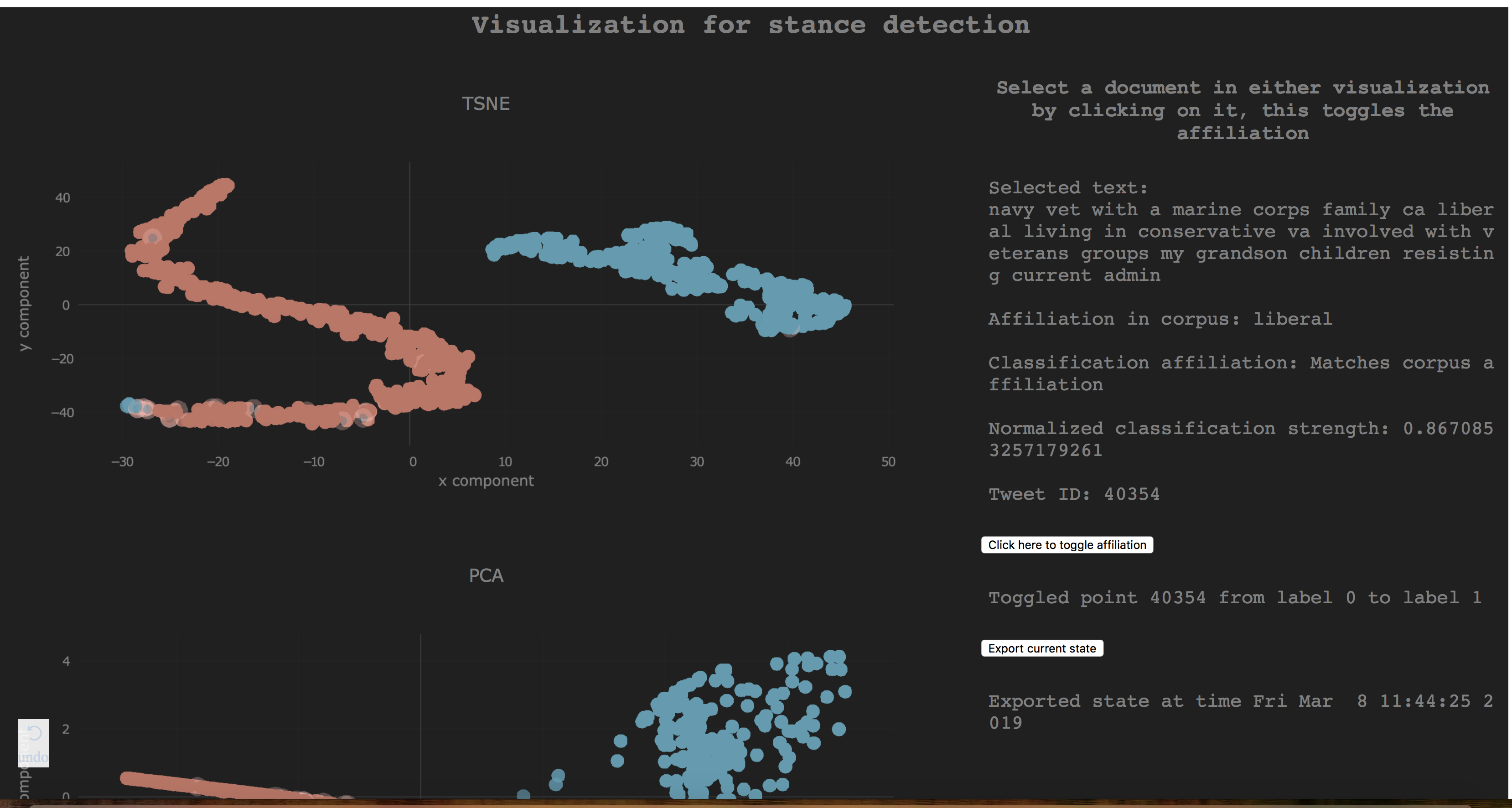}
    \caption{Framework for stance detection}
    \label{application}
\end{figure}

A web-based interactive application (\autoref{application}) is created to streamline this process of decision making. While tests were conducted with a single user, the same process can also be performed collaboratively with a group of users to leverage expertise from multiple fields. While a lot of trained models produce good results they rarely provide interpretability, i.e. the user is never informed how the model made those decisions. The suitability of various pretrained static and contextual embeddings are evaluated for this purpose. An Attention layer and the weights it outputs provide a level of transparency in this decision-making process. We summarize the performance of these models and also outline the hyperparameters that have an impact quantitatively on classification performance and qualitatively on solution interpretability.

To summarize, the goal of this paper is to assess the feasibility of a weakly supervised DNN to produce projections that requires minimal human supervision to correct labelling errors as opposed to manually and laboriously inspecting the entire corpus. In this regard, the key contributions of this paper are as follows:

\begin{itemize}
\item  Evaluation of Deep Neural Network (DNN) configurations for Natural Language Understanding (NLU) trained to evaluate political affiliation. Various static and contextual embeddings are evaluated here.
\item  Evaluation of various dimensionality reduction techniques for visualization.
\item  The creation of a web-based interactive application that integrates the results of the DNNs through 2D projections which allows for human-in-the-loop decision making to classify documents in a corpus.
\end{itemize}

\section{Methodology}

\subsection{Corpus generation and preprocessing}

\begin{figure}[tb]
  \centering
  \includegraphics[width=\columnwidth]{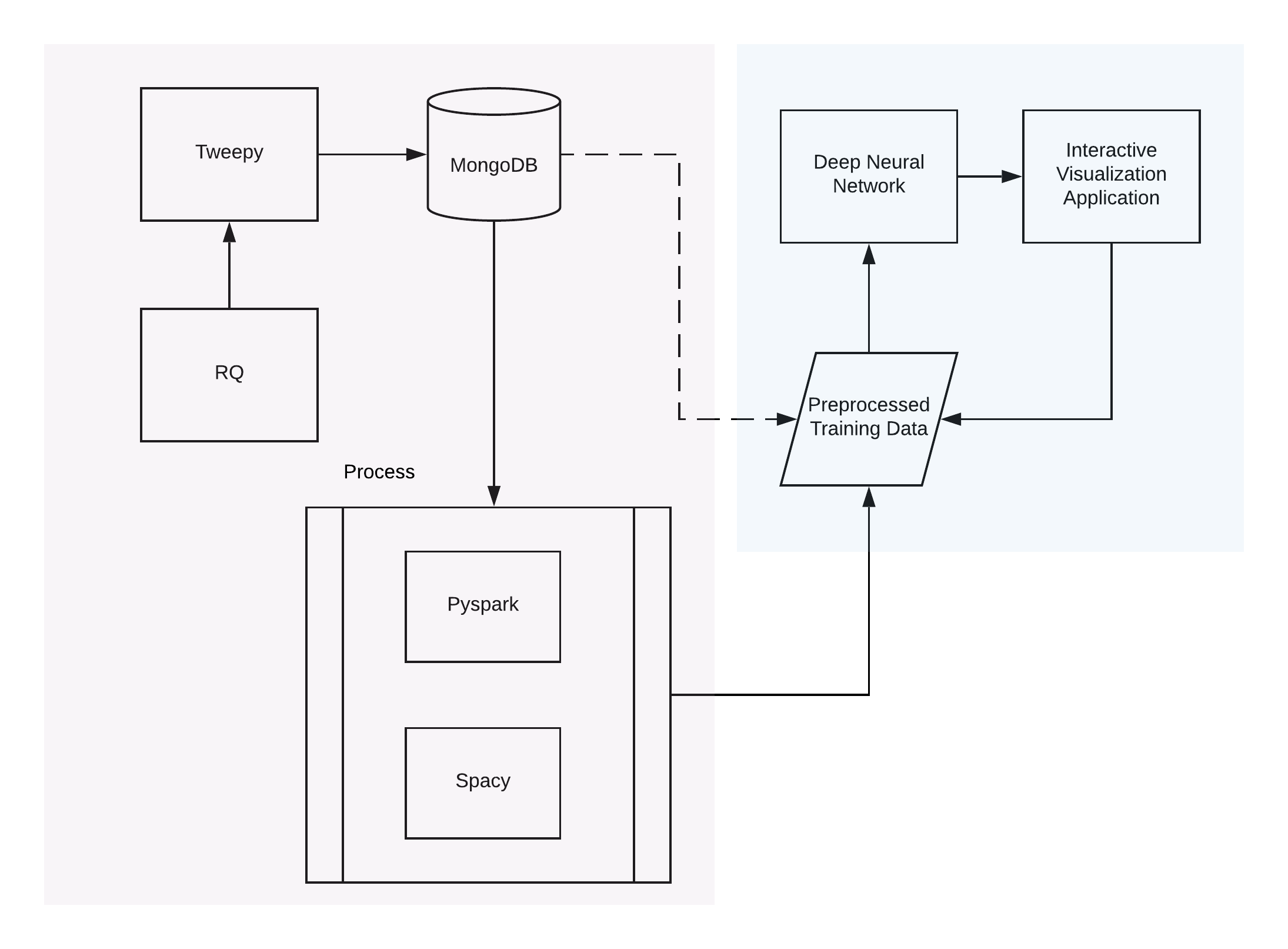}
  \caption{Overview of the NLP pipeline}
  \label{overview}
\end{figure}

A high-level overview of the pipeline used to used to procure and process the data is shown in \autoref{overview}. The block in blue is the part of the pipeline with which the user interacts. The corpus was created by downloading tweets from Twitter corresponding to the political affiliations of 'conservative' and 'liberal' users. This was done with a domain expert who has the knowledge to guide the learning algorithm. These tweets along with the metadata are stored in a NoSQL MongoDB \cite{mongodb} database on a server. As a result of the weak supervision, not only are the labels noisy but the text in the corpus is also noisy due the nature of the source. The corpus is preprocessed using the NLTK library \cite{nltk} and the PyMongo library's regular expression parsing. No stemming or lemmatization was done, however normalization and cleaning was done to reduce the amount of noise in the dataset. The data was then fed, using the TorchText library \cite{torchtext}, to a network created in PyTorch to train such that the weights learn to differentiate the two distinct ideologies. This process is distinctly different from sentiment analysis in the sense that one has to learn an entity's opinion towards a number of topics that constitute a person's political affiliation here, as opposed to simply expressing a positive or negative sentiment. In a way, this can be seen as a type of target-based sentiment analyses also known as Aspect-Based Sentiment Analysis (ABSA) \cite{absa}, where the target topics are continually updated.

The data was downloaded periodically using the job scheduler RQ \cite{rq} for three months. The downloader was written using the Python library Tweepy \cite{tweepy}. Although around 2.6 million tweets were downloaded over this period, after eliminating users with either no TUDs or non-descriptive TUDs and discarding duplicates, there were 78075 conservative and 49116 liberal unique TUDs. Data input to the DNN has the TUD, label and tweet id associated with it so that the end-user can cross-reference that with the database. A baseline accuracy is estimated using the remaining for the unbalanced test set to measure performance improvements.

\subsection{Overview of Network Architecture}

\begin{figure}[h]
  \centering
  \includegraphics[width=\linewidth]{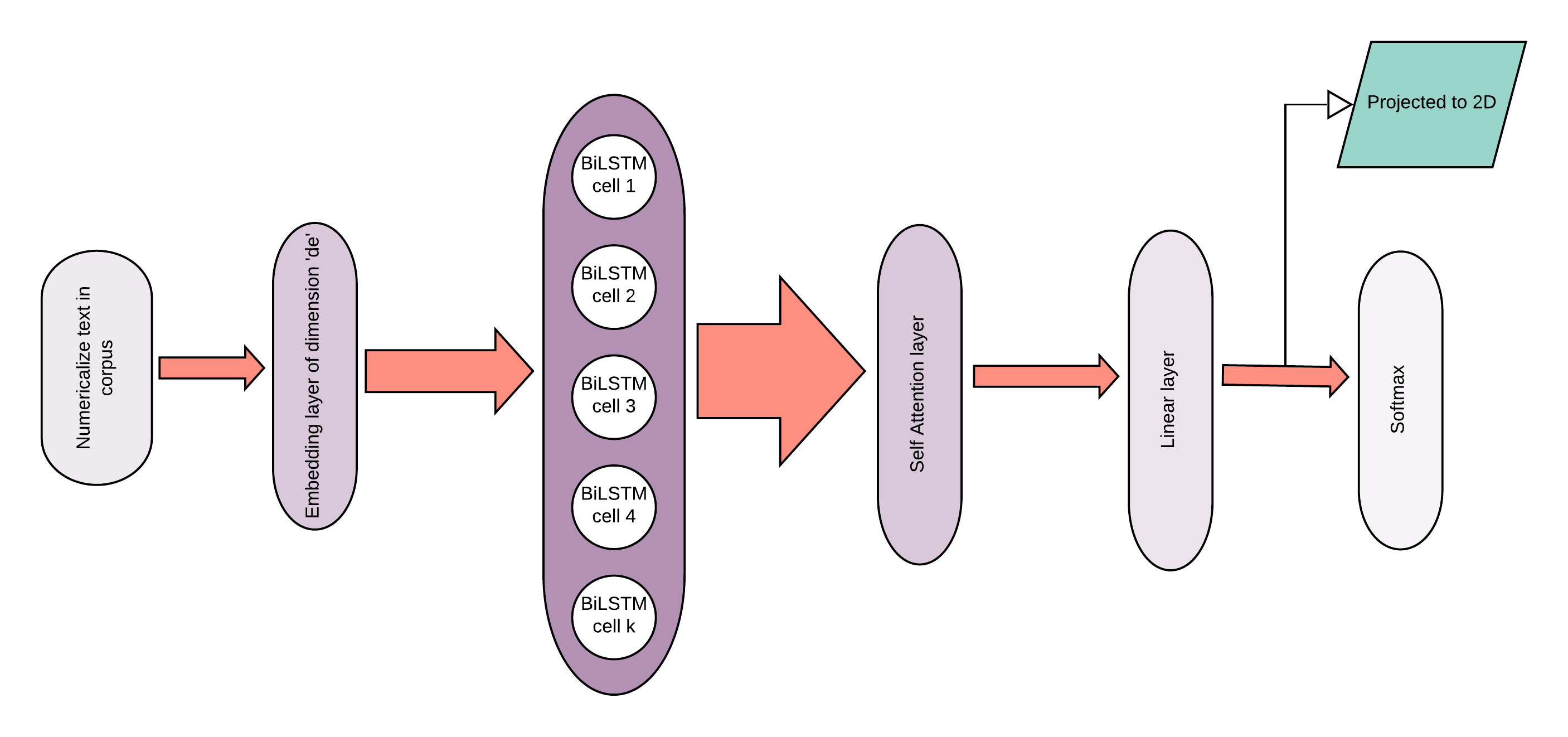}
  \caption{Architecture of the Neural Network; the layer color intensities indicate computational needs and arrows indicate magnitude of data flow.}
  %\Description{Architecture}
  \label{fig:network}
\end{figure}

An overview of the network used in the classification and feature extraction is shown in  \autoref{fig:network}. The network consists of the following layers in order: embedding layer, bi-directional LSTM, self-attention layer \cite{selfattention,attention-paper}, dense layer, and a softmax layer. The direction and magnitude of data flow between the layers is indicated by the arrows connecting them (\autoref{fig:network}). Initially, the embedding layer is not loaded with pretrained embeddings so that it can be tailored to the specific problem posed here. The experiment is repeated with the embedding layer preloaded with the 100-dimensional Glove \cite{glove} embeddings. Sequence length is variable corresponding to the maximum sequence length of the sentences in a batch during the training or evaluation pass. The shorter sentences are padded as needed. Depending on the distribution of sequence lengths in the corpus this could potentially be an inefficient process, affecting both speed and accuracy of the model. To present sequences of varying length to an LSTM, one can use a `packed' sequence such that the LSTM does not see a padding token, thereby resulting in potential computational cost savings.

 The dimensionality of the bidirectional LSTM layer is `n', resulting in a size of `2n' for both the output and hidden states corresponding to the forward and backward passes. The embedding layer generates word-level embeddings which are then passed on to the bidirectional LSTM which is notable for being able to learn from sequential data. Batch size was set at 512 although the upper bounds of both the batch size and the LSTM dimensionality depends on the amount of GPU memory available on the system. Although in most situations involving an LSTM only the output is required downstream, the use of an attention layer requires the full hidden state of the LSTM. As a result, the hidden states are connected to the self-attention layer which generates an attention vector and attention weights corresponding to each token in the sentence. The generated  attention vector is passed downstream through the network while the attention weights are used to visualize word relevance at the sentence level as shown in \autoref{attention}. A more detailed explanation of the attention mechanism follows in the section `Model Interpretability'. The output of the penultimate layer (dense layer) is chosen for projection and visualization (\autoref{application}).

%Did you use pretrained embeddings, if so which one - provide accuracy improvements in results. Explain the reason for bidirectional LSTM. Show charts show how accuracy changes with LSTM hidden layer size. Explain how sequence length varies in the corpus, maybe a histogram of word length in corpus - used fixed word size of 20 at first in keras, changed to max length of batch in Pytorch - explain why packed layers are important and show how accuracy increased with the use of packed layers. What is a self-attention layer - Why is it necessary - to make the results interpretable by visualizing the weights. 

\subsection{Interactive and exploratory web-based application}

The interactive web-based application (\autoref{application}) that is created can be divided into two halves: the left half presents the projections and the right half presents the information associated with an entity selection. To interact with the visualization the user can single click on an entity in the visualization to select it, or the user can use any of the pan/zoom/select tools to explore the visualizations. The top projection is the result of the application of t-SNE to the output of the penultimate layer and the bottom projection is the MDS projection of the same output with euclidean distance functions.

% \begin{figure}[h]
% \begin{minipage}[t]{0.45\textwidth}
% \includegraphics[width=\linewidth]{pca_orth.png}
% \caption{Structure of PCA projections from the penultimate layer}
% \label{pca_orth}
% \end{minipage}
% \hfill
% \begin{minipage}[t]{0.45\textwidth}
% \includegraphics[width=\linewidth]{tsne_out.png}
% \caption{Structure of TSNE projections from the penultimate layer}
% \label{tsne_out}
% \end{minipage}
% \end{figure}

\begin{figure}[h]
  \centering
   \includegraphics[width=\linewidth]{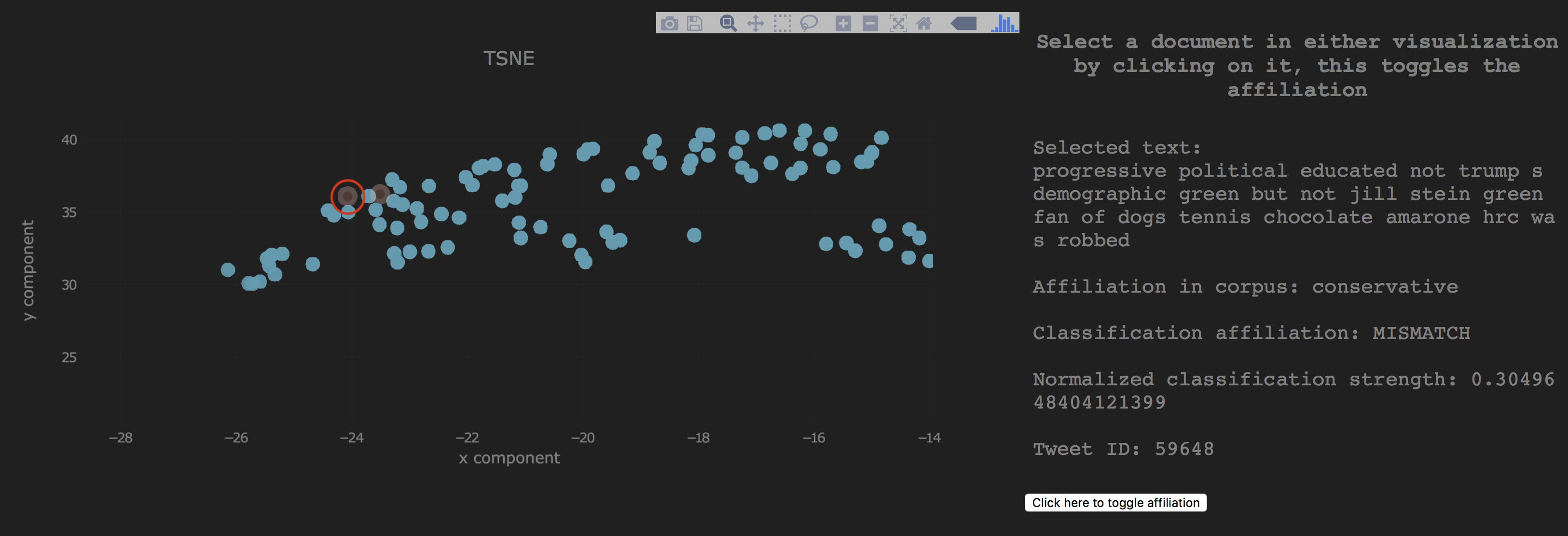}
   \caption{Example of a type `a' error with an entity incorrectly labeled as 'conservative' }
   %\Description{lib}
  \label{lib}
\end{figure}

\begin{figure}[h]
  \centering
   \includegraphics[width=\linewidth]{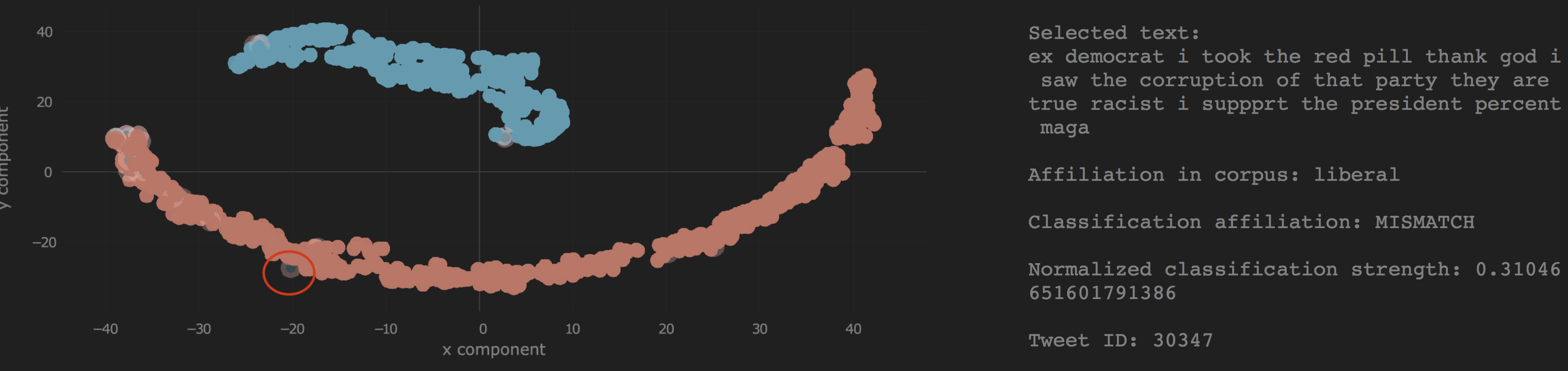}
   \caption{Example of a type `a' error with an entity incorrectly labeled as 'liberal' }
   %\Description{cons}
  \label{cons}
\end{figure}

\begin{figure}[h]
  \centering
   \includegraphics[width=\linewidth]{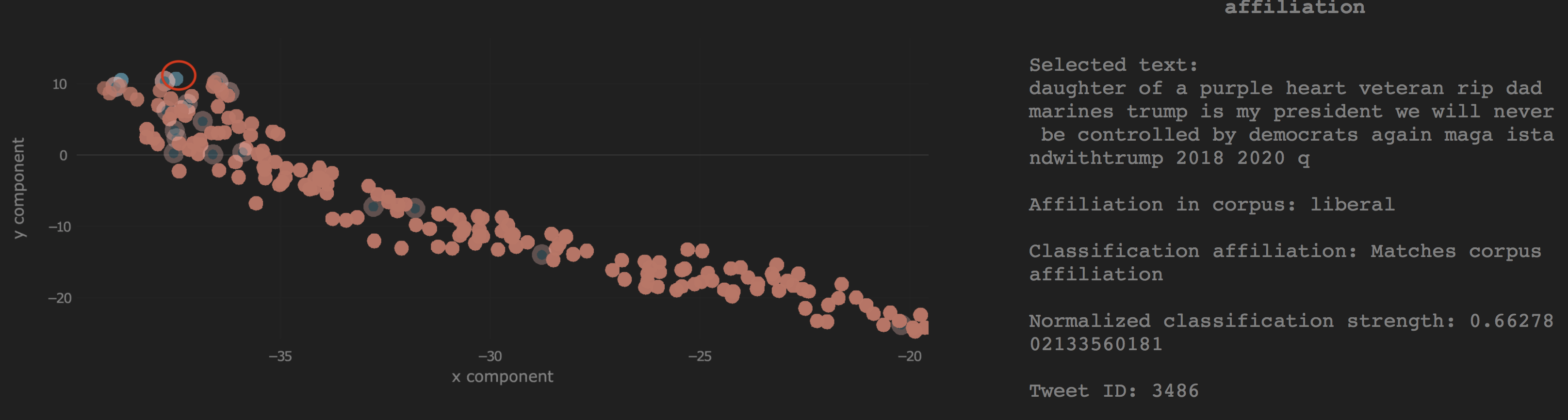}
   \caption{Example of an outlier that is a type `a' error}
   %\Description{outlier}
  \label{outlier}
\end{figure}

 Entities are color-coded to represent liberal (blue) and conservative (red) ideologies. Distinct clusters are visible in both projections, however there are some entities that are either inaccurately labeled due to the weak supervision (type `a' or aleatoric uncertainty) or incorrectly classified by the DNN (type `b' or epistemic uncertainty). We seek to identify the ones that are inaccurately labeled, i.e. reduce type `a' error, and correct them such that the impurity in the corpus label decreases. Note that this should aid in the eventual reduction of type `b' errors as well. Examples of type `a' errors are shown in \autoref{cons} and \autoref{lib}. These projections also have the added benefit that the proximity in the two-dimensional space can be used to represent relative 'political affiliation' without having to quantify it subjectively.

The projections are color and opacity coded for affiliation in corpus, and strength of classification respectively. Entities with a high normalized classification strength score are more opaque indicating more confidence in the prediction while those with a lower score (closer to 0.5) are more transparent. Currently, the normalized classification strength is calculated from the softmax score for an entity although other techniques are currently being evaluated for uncertainty quantification. A ring around incorrect classifications alerts the user to that particular entity; it is marked for inspection so that the user can determine the type of error (type a or b). The projections illustrate distinct clustering in the two categories. Often, there are outliers that do not fit into their respective clusters as can be seen by the blue glyphs interspersed with the red cluster in \autoref{application} and \autoref{outlier}, these are also candidates for inspection. These two types comprise the `entities under review'. The user can review these entities and correct them if needed, and once this is done the corrected corpus can be exported to the training corpus for iterative retraining of a model and corpus cleaning. This allows for the user to efficiently perform iterative cleaning of the corpus by only focusing on the `entities under review' thereby significantly expediting the process.

\subsection{Model interpretability}

With the prevalence and success of the predictive power of DNNs, it has also faced criticisms over how the results were generated. This has accelerated efforts to provide a solution to this concern, which is informally referred to as Interpretable AI. While the application shown in \autoref{application} allows us to assess stance with a measure of uncertainty, how this determination was made is not transparent to the user. 

\begin{figure}[ht]
  \centering
   \includegraphics[width=0.9\linewidth]{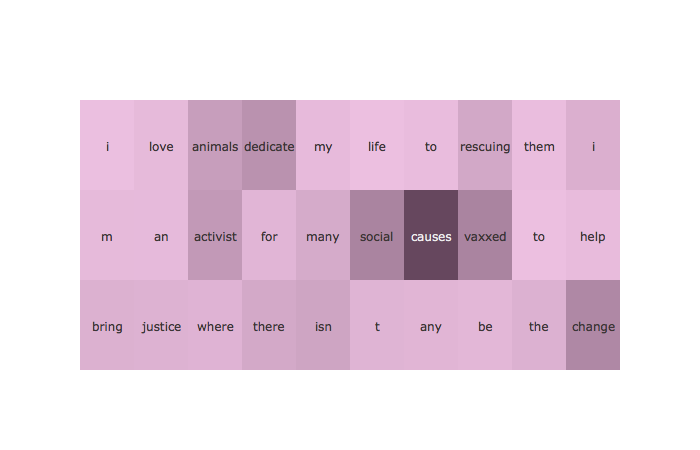}
   \caption{Illustration of Attention weights for model interpretability}
   %\Description{attention}
  \label{attention}
\end{figure}

An attention layer takes as input a `context' and `query' and computes the similarity of the query vector to each vector in the context matrix. In self-attention, the context and query are the same and one computes the similarity of each word in the sequence to every other word in this sequence to form an attention weight matrix. Summing up the weights across all the rows produces an attention weight vector.The self-attention layer that follows the LSTM layer outputs a set of attention weights and an attention vector which is the weighted sum of the LSTM hidden states and the attention weights. The attention vector thus formed is fed downstream to the dense linear layer while the attention weights can be used to visualize the relevance of each individual word in a sentence to its classification. An example of this is illustrated in \autoref{attention}. The emphasized words (darker boxes) have a larger contribution to the classification score or outcome, thereby informing the user what words are relevant from the network's perspective. The attention weights are normalized over all the words in a sentence, excluding any padding words, so they can be visualized.

\begin{figure}[ht]
  \centering
   \includegraphics[width=0.9\linewidth]{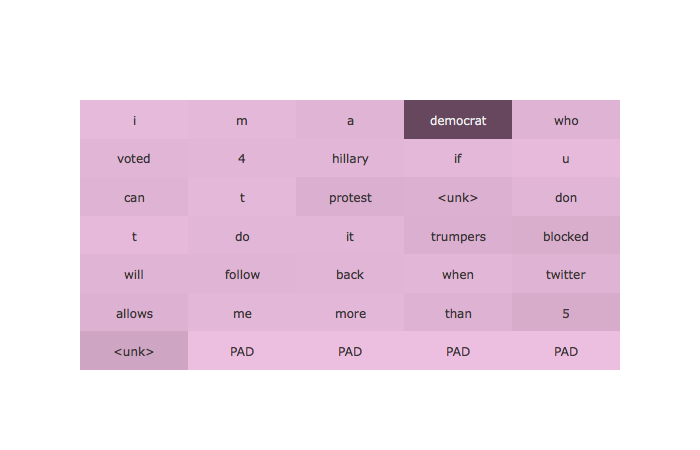}
   \caption{Attention weights with unknown tokens}
   %\Description{attention}
  \label{attention2}
\end{figure}

\subsection{Elmo: Using contextual embeddings}

In this section we look at replacing statically pretrained embeddings with contextual embeddings, namely Elmo \cite{Elmo}. Elmo is a type of deep contextualized word embedding that can more accurately model the syntax and semantics of usage within a sentence. The advantage of Elmo over static embeddings is the ability to model polysemy, or variations of word usage across contexts. These contextual embeddings themselves are generated using a Bidirectional Language Model which has been trained on an extensive corpus.

\section{Results}
% 99.12 for size 4 hidden dim, 99.23 for 8, 99.33 for 16, 99.39 for 32, 99.37 for 64, 99.37 for 128, 99.38 for 256
% Number of unknowns and vocab size, numbers from test set - 23829 for 25000, 12857 for 50000 words, 5661 for 75000 words, 3800 for 100000 total words in test is 538379

\subsection{Corpus statistics}

\begin{figure}[ht]
\begin{minipage}[t]{0.45\textwidth}
\includegraphics[width=\linewidth]{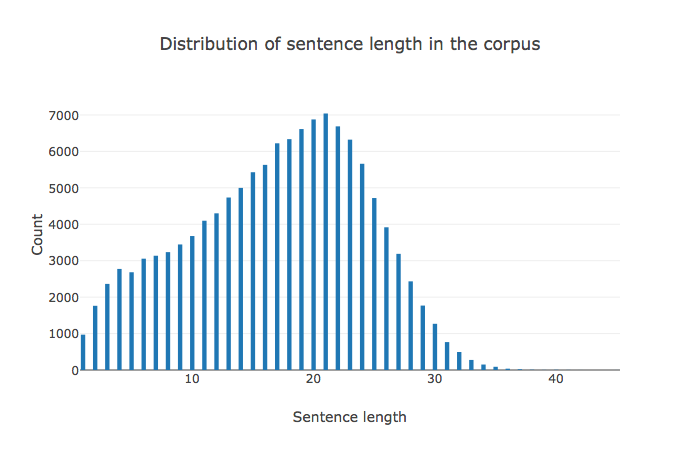}
\caption{Distribution of sentence lengths in the corpus}
\label{distribution}
\end{minipage}
\hfill
\begin{minipage}[t]{0.45\textwidth}
\includegraphics[width=\linewidth]{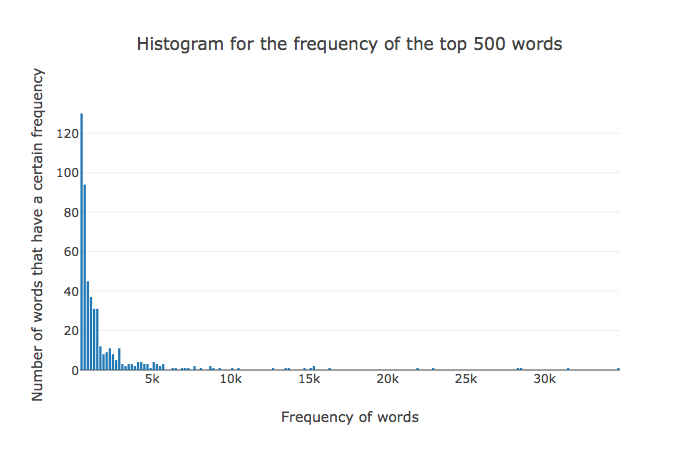}
\caption{Count of word frequencies in the corpus}
\label{word_freq}
\end{minipage}
\end{figure}

\begin{figure}
\begin{minipage}[t]{0.45\textwidth}
\includegraphics[width=\linewidth]{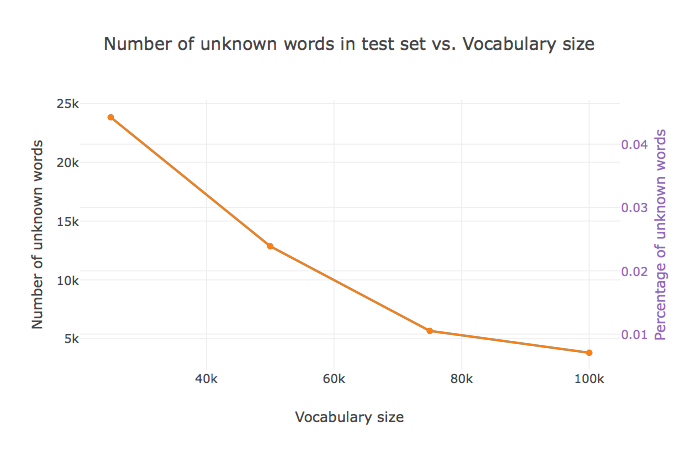}\caption{Number of unknown words in test set compared to the vocabulary size} 
\label{vocabulary}
\end{minipage}
\hfill
\end{figure}

% \begin{figure}[h]
%   \centering
%    \includegraphics[width=\linewidth]{sentence_distribution.png}
%    \caption{Distribution of sentence lengths in the corpus}
%    \Description{distribution}
%   \label{distribution}
% \end{figure}

% \begin{figure}[ht]
%   \centering
%    \includegraphics[width=\linewidth]{word_freq.png}
%    \caption{Count of word frequencies in the corpus}
%    \Description{word_freq}
%   \label{word_freq}
% \end{figure}

% \begin{figure}[ht]
%   \centering
%    \includegraphics[width=\linewidth]{vocab.png}
%    \caption{Number of unknown words in test set compared to the vocabulary size}
%    \Description{vocabulary}
%   \label{vocabulary}
% \end{figure}

% \begin{figure}[ht]
%   \centering
%    \includegraphics[width=\linewidth]{accuracy.png}
%    \caption{Accuracy of Classifier vs. LSTM dimension size}
%    \Description{accuracy}
%   \label{accuracy}
% \end{figure}

To get a sense of the corpus that is being used here, the distribution of the sentence lengths are plotted in \autoref{distribution}. It can be seen from the figure why it is benefical to use a variable length sequence. The number of words in the vocabulary is a hyperparameter that has an impact on how well the model performs; it has a greater impact on model interpretability through attention weights than classification performance. A chart that displays how the number of unknown words in the test set varies with the vocabulary size is shown in \autoref{vocabulary}. A histogram of the frequency of the top 500 words in the corpus is shown in  \autoref{word_freq}. An example of a visualization with the vocabulary size set to 10000 is shown in \autoref{attention2}; a larger number of 'unknown' tokens now show up here thereby impeding the performance of the attention layer and thereby the model's interpretability.

\subsection{Model comparison}

\begin{figure}[!h]
\begin{minipage}[t]{0.45\textwidth}
\includegraphics[width=\linewidth]{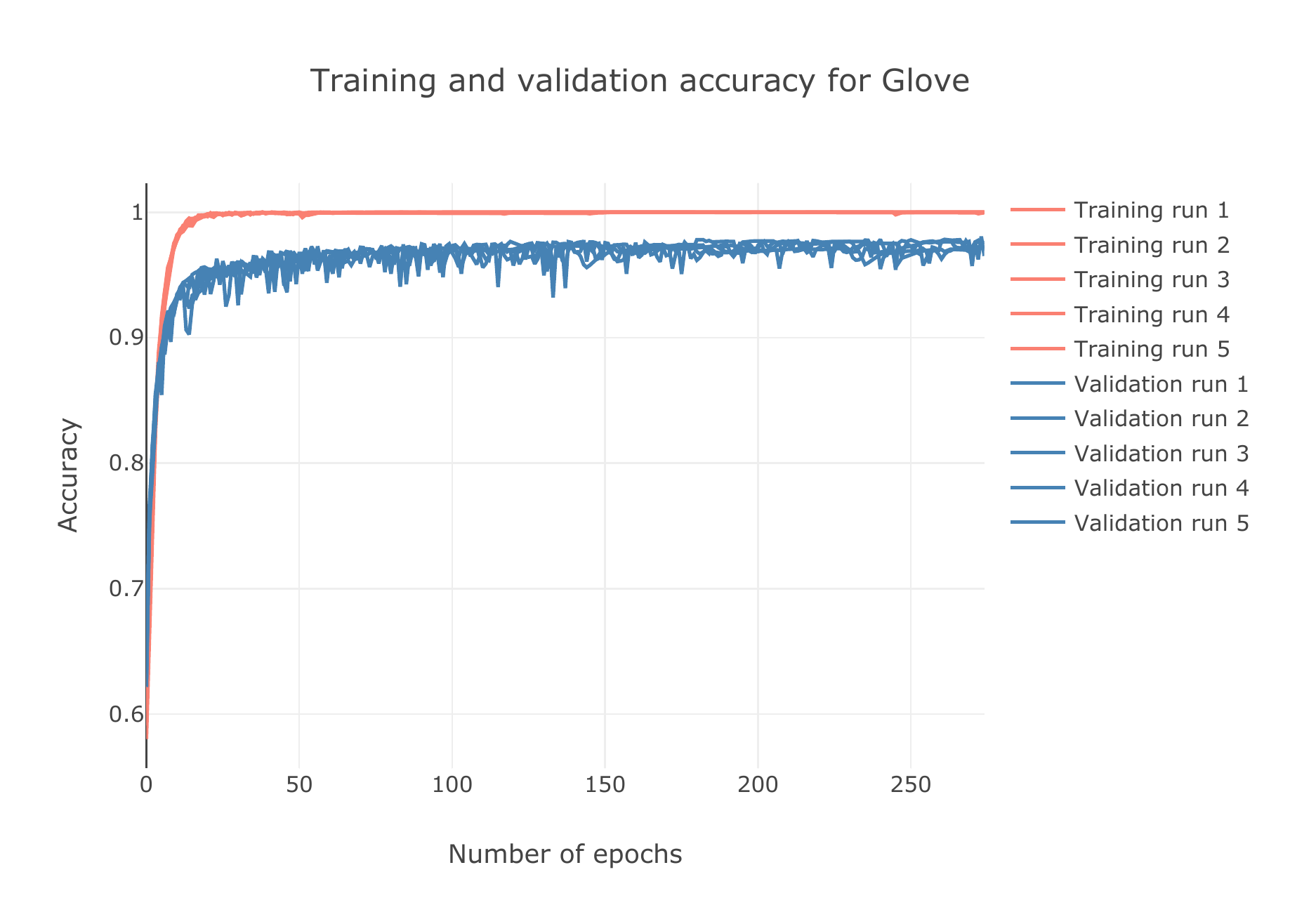}
\caption{History of training and validation accuracy for Glove}
\label{glove_accuracy}
\end{minipage}
\end{figure}

\begin{figure}[!h]
\begin{minipage}[t]{0.45\textwidth}
\includegraphics[width=\linewidth]{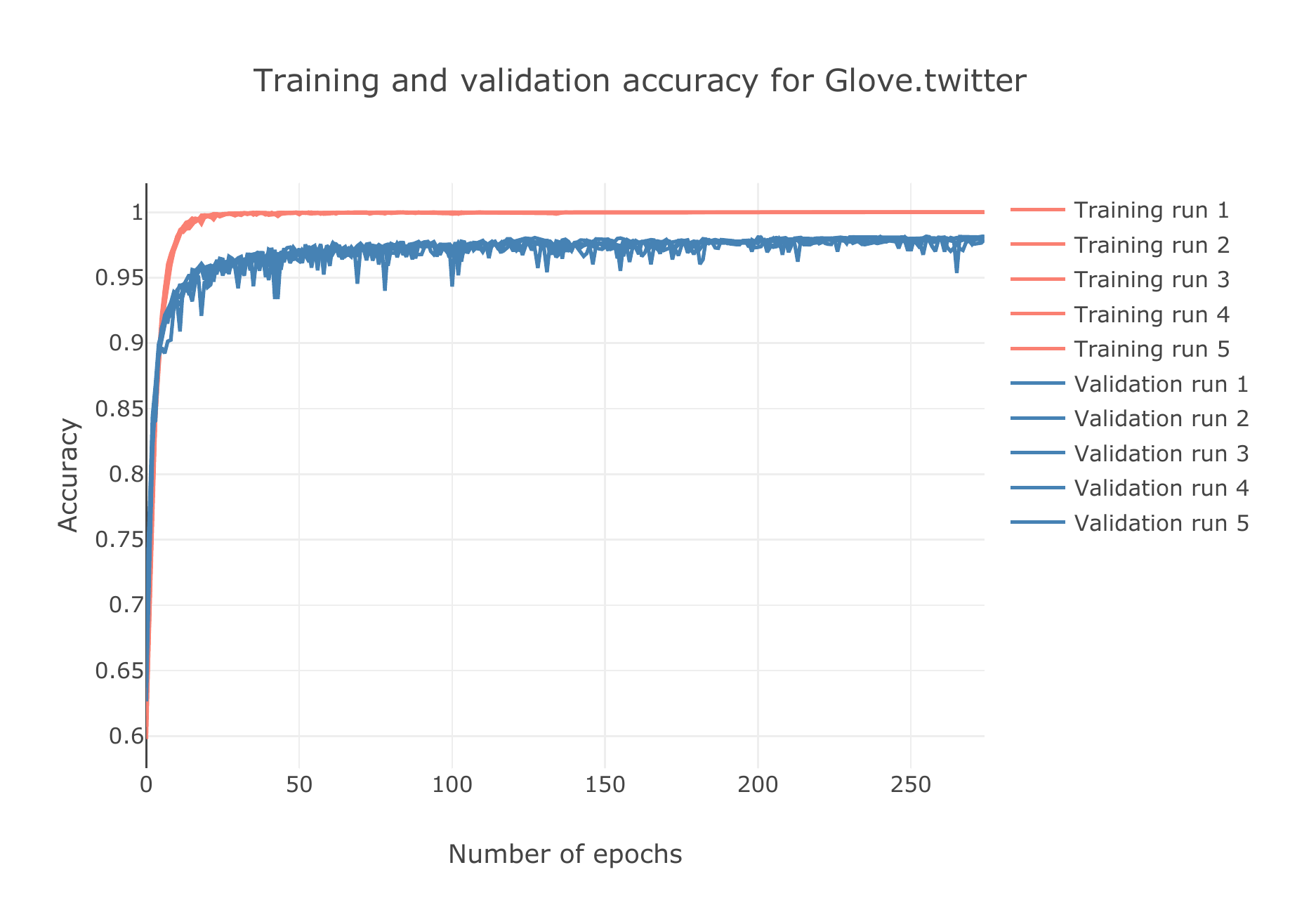}
\caption{History of training and validation accuracy for Glove.twitter}
\label{glove_twitter_accuracy}
\end{minipage}
\end{figure}

\begin{figure}[!h]
\begin{minipage}[t]{0.45\textwidth}
\includegraphics[width=\linewidth]{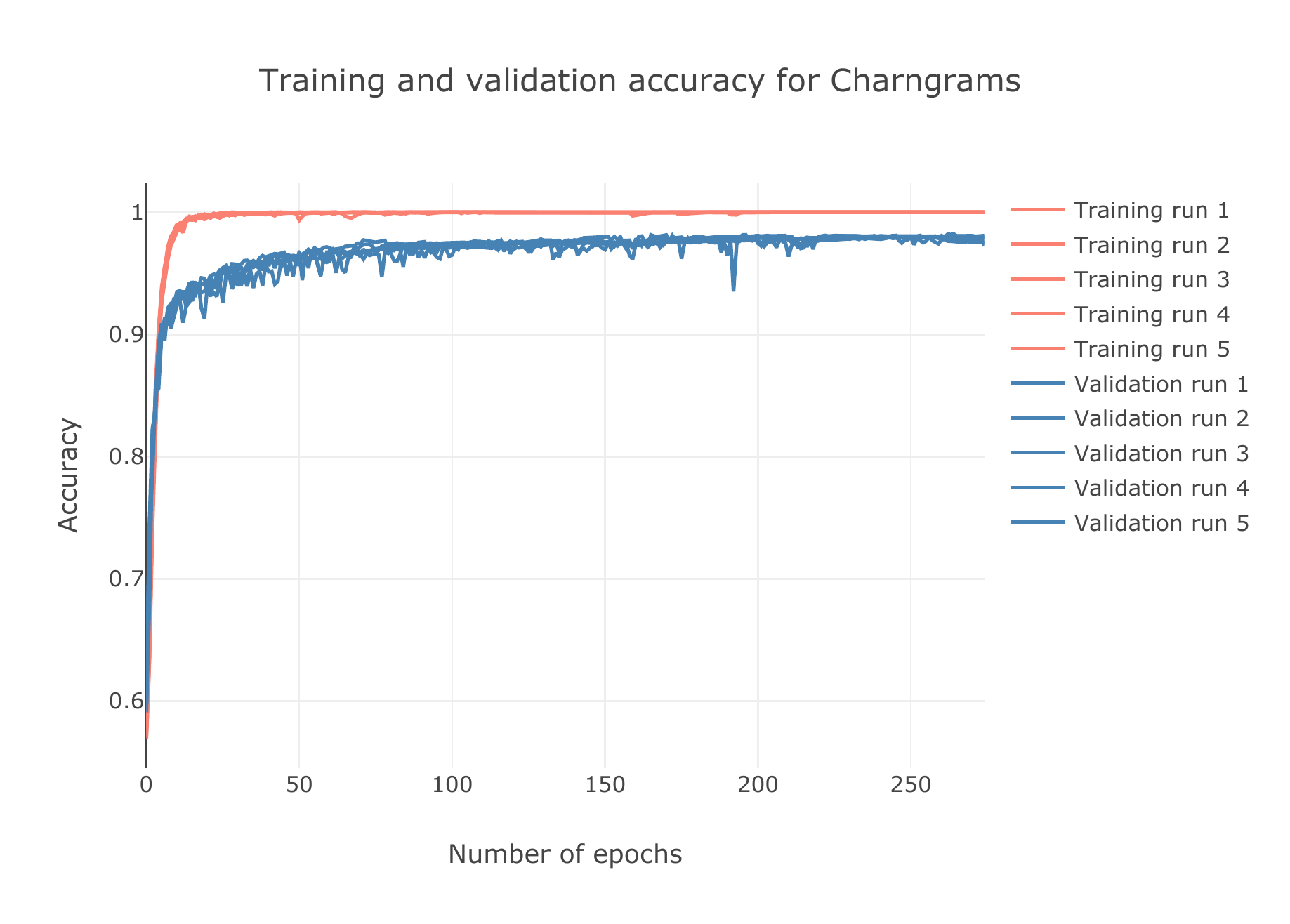}
\caption{History of training and validation accuracy for Charngrams}
\label{charngrams_accuracy}
\end{minipage}
\end{figure}

\begin{figure}[!h]
\begin{minipage}[t]{0.45\textwidth}
\includegraphics[width=\linewidth]{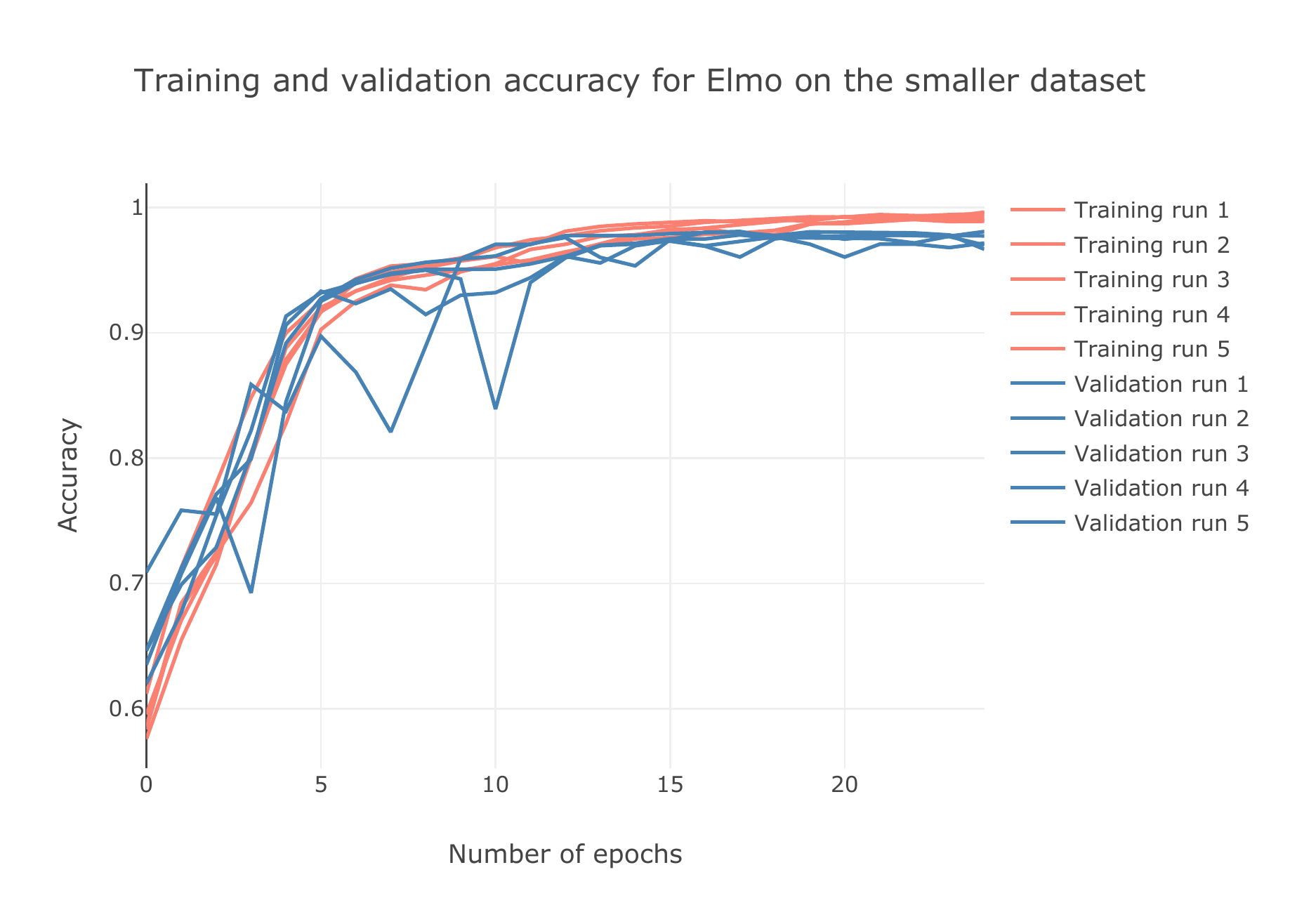}
\caption{History of training and validation accuracy for Elmo on small corpus}
\label{elmo_small_accuracy}
\end{minipage}
\end{figure}

\begin{figure}[!h]
\begin{minipage}[t]{0.45\textwidth}
\includegraphics[width=\linewidth]{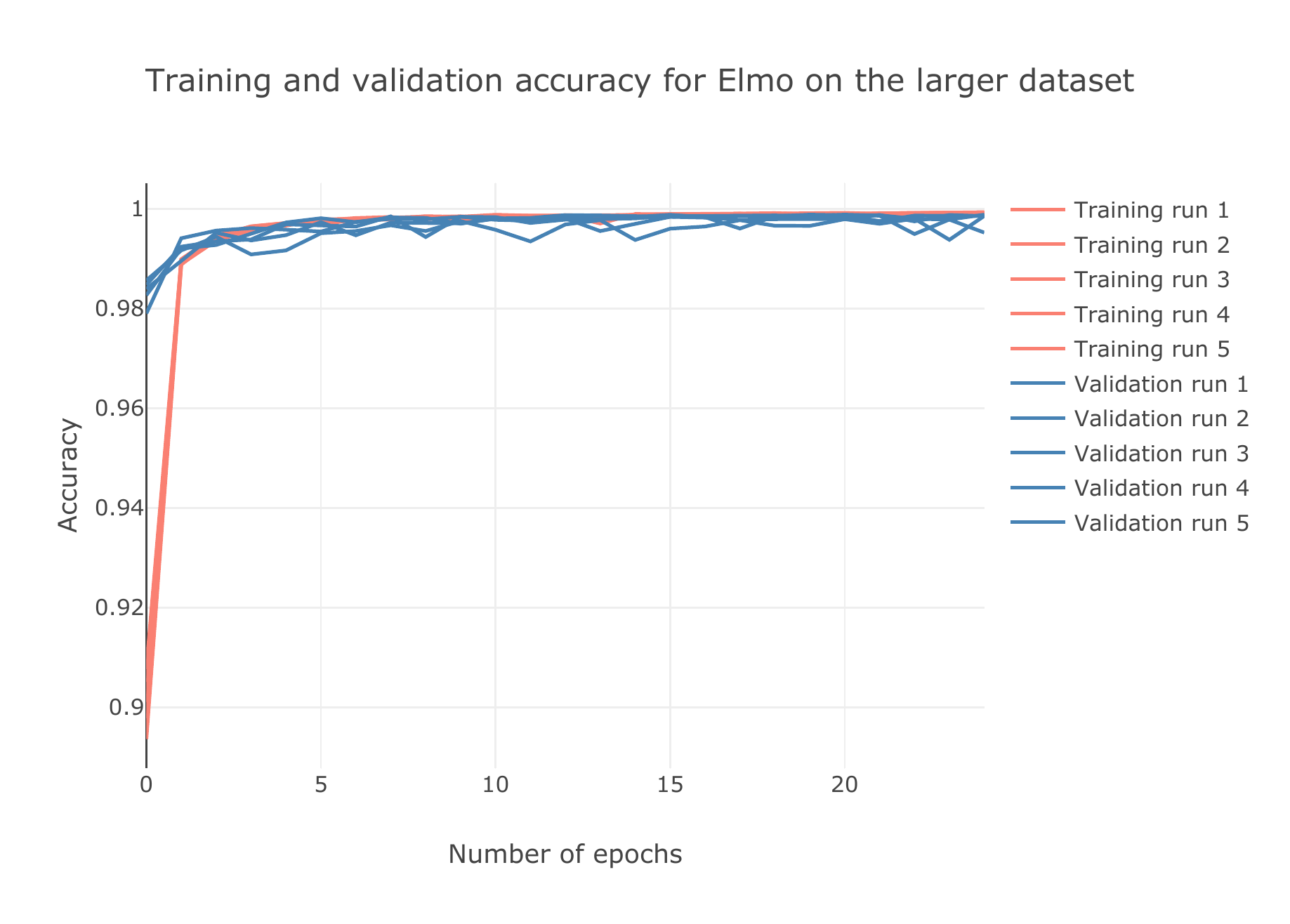}
\caption{History of training and validation accuracy for Elmo on full corpus}
\label{elmo_large_accuracy}
\end{minipage}
\end{figure}

\begin{figure}[!h]
\begin{minipage}[t]{0.45\textwidth}
\includegraphics[width=\linewidth]{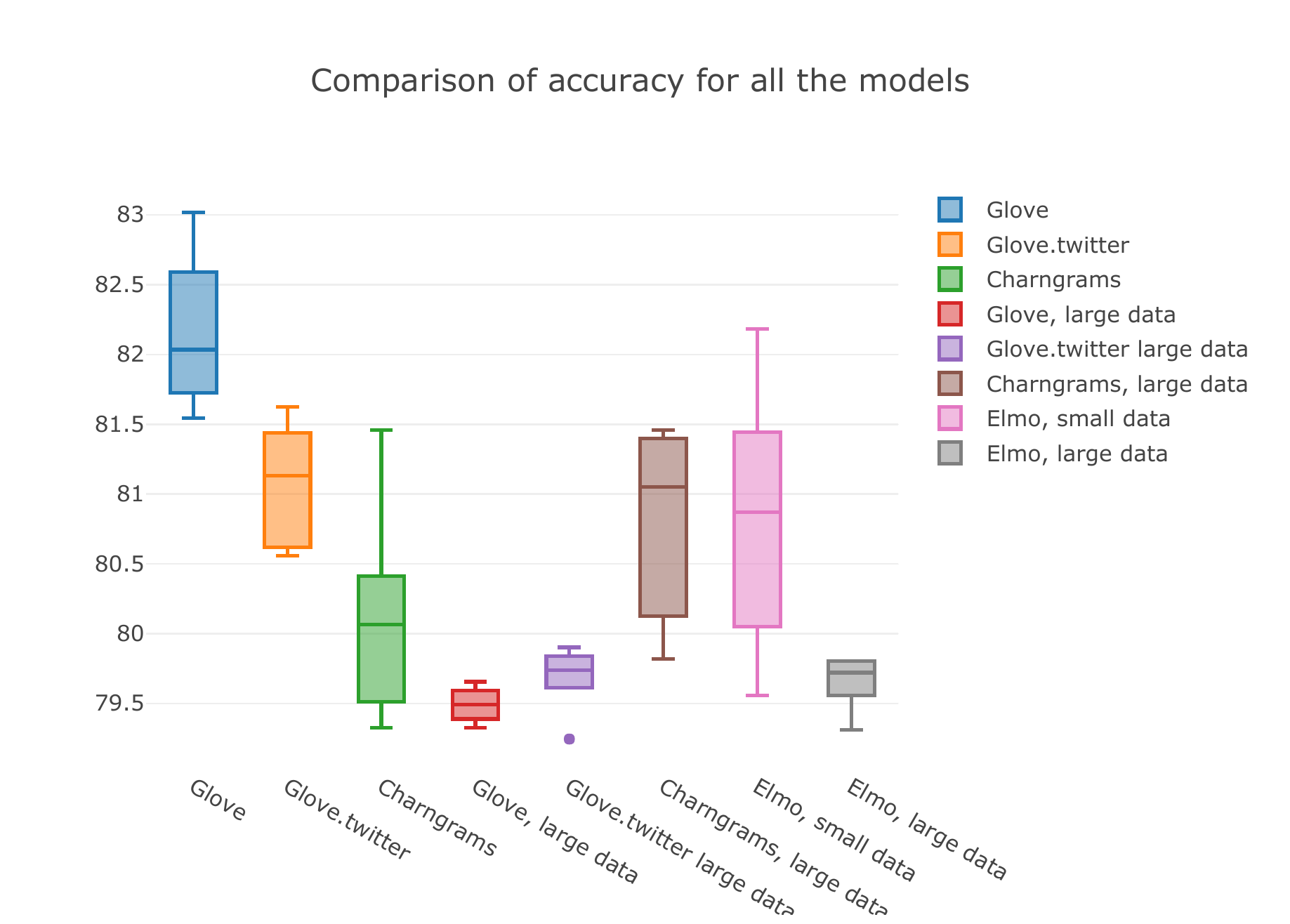}
\caption{Comparison of accuracy}
\label{comparison_accuracy}
\end{minipage}
\end{figure}

\begin{figure}[!h]
\begin{minipage}[t]{0.45\textwidth}
\includegraphics[width=\linewidth]{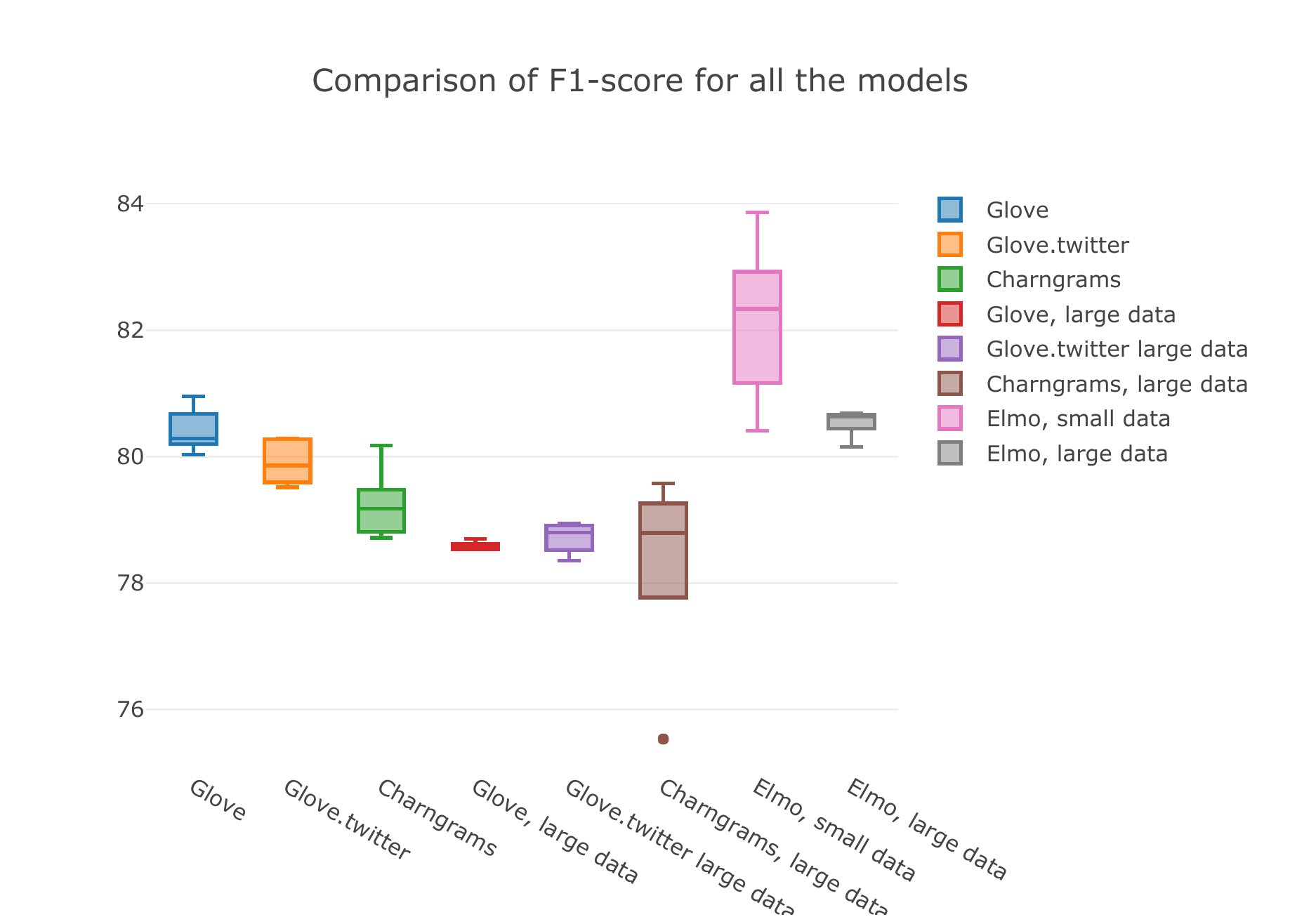}
\caption{Comparison of F1-scores}
\label{comparison_f1}
\end{minipage}
\end{figure}

\begin{figure}[!h]
\begin{minipage}[t]{0.45\textwidth}
\includegraphics[width=\linewidth]{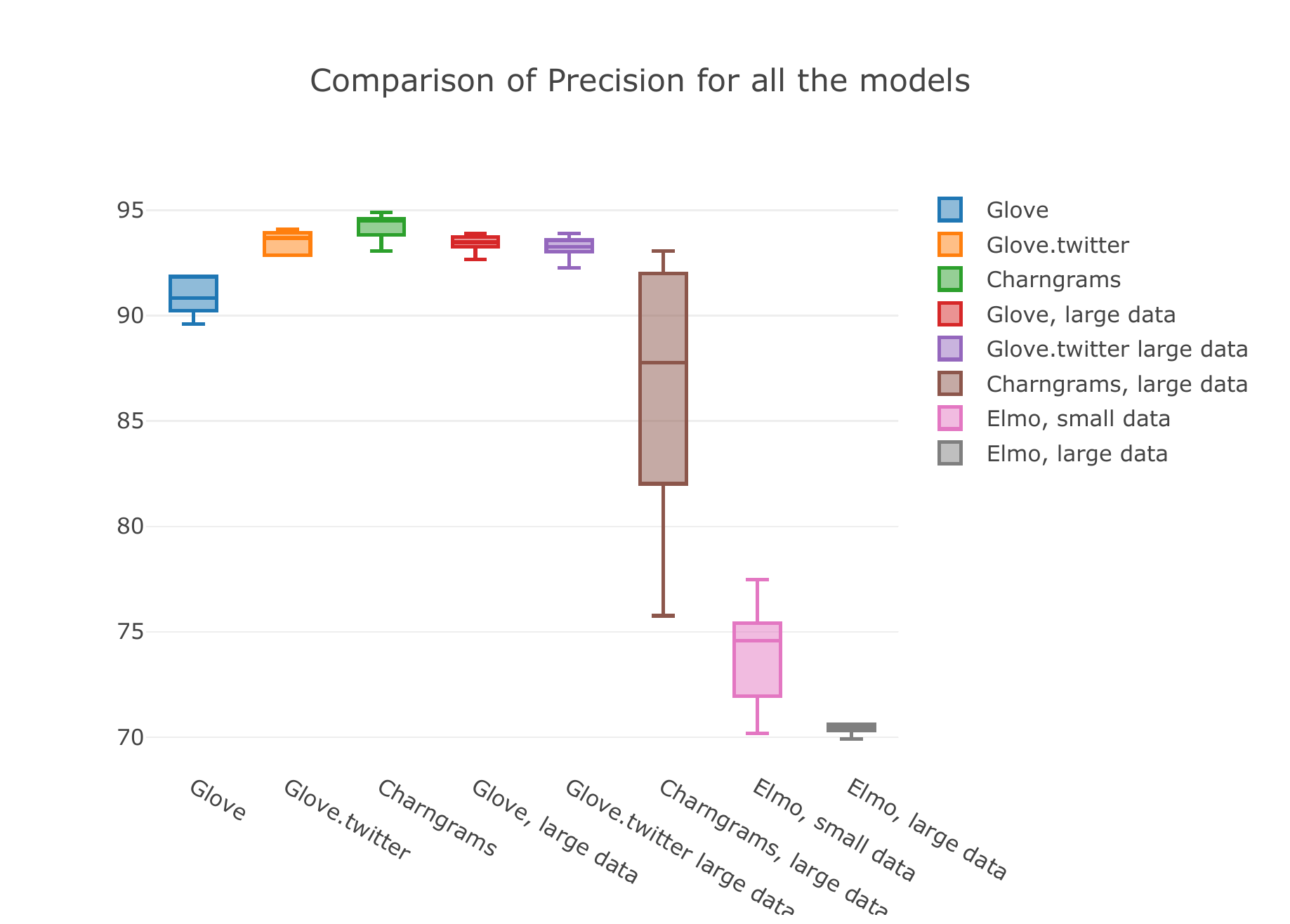}
\caption{Comparison of precision}
\label{comparison_precision}
\end{minipage}
\end{figure}

\begin{figure}[!h]
\begin{minipage}[t]{0.45\textwidth}
\includegraphics[width=\linewidth]{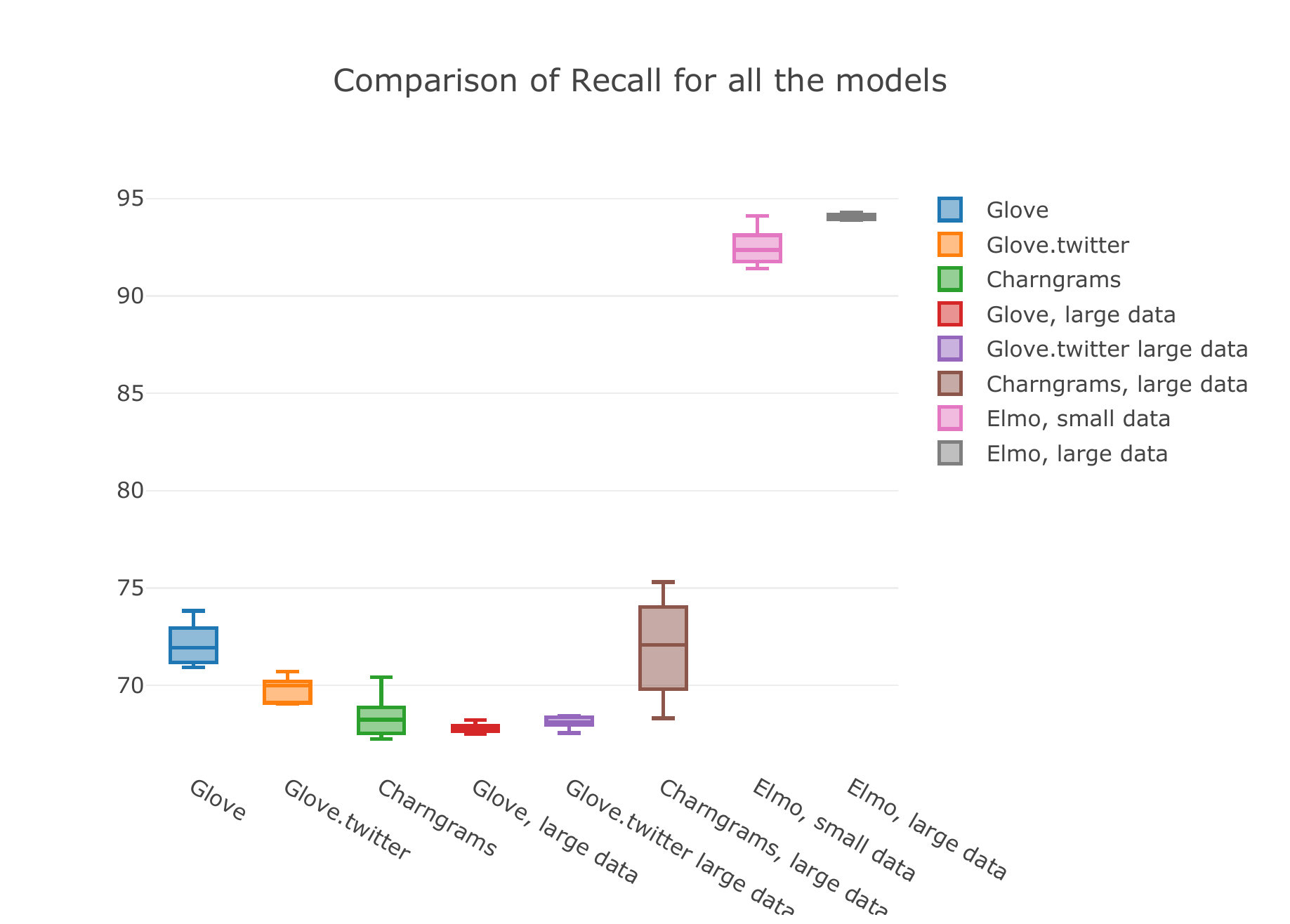}
\caption{Comparison of recall}
\label{comparison_recall}
\end{minipage}
\end{figure}

\begin{figure}[!h]
\includegraphics[width=\linewidth]{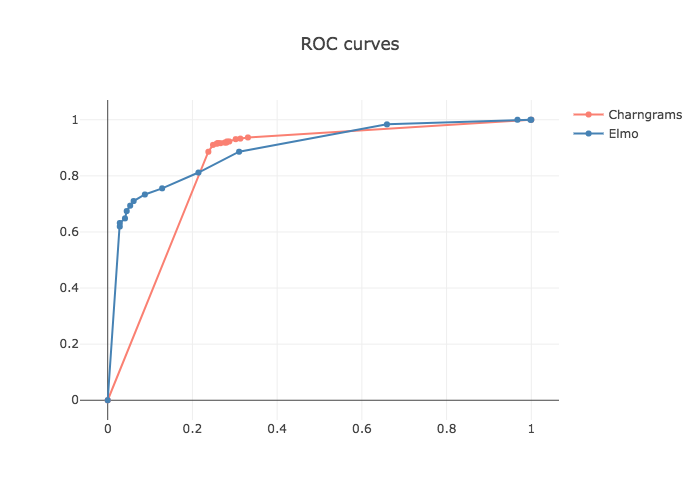}
\caption{ROC curves for BiLSTM with charngrams embeddings and Elmo}
\label{roc_curves}
\end{figure}

Performance of the network is evaluated in terms of prediction accuracy and efficacy of the visualizations produced. All models were run for 25 epochs to get a measure of prediction accuracy. Each model, in turn, was trained/validated and tested on the weakly-supervised data and the annotated test data respectively 5 times to account for model variance. The following measurements were computed with a vocabulary size of 75000 words. Originally, a fixed word size of 20 was used for the sentence length which was later changed to a fixed length corresponding to the longest sentence in the batch and then to a packed sequence which eliminates unnecessary padding propagating through the LSTM layer. Single iteration trials received an average increase in validation accuracy of 1.95\%. Changing to a variable length sentence using packed LSTM layers further boosted the average accuracy by 0.29\%. Hyperparameter tuning was done manually for the BiLSTM hidden dimension size, the size of the forward linear layer and the dropout parameter. A grid search or another AutoML-based approach would probably yield better results than what is possible with manual hyperparameter tuning.

There were two sets of training data generated from the entire corpus of 127191 records. The first one utilized the entire data and the second one was a smaller dataset intended to assess performance of the network configurations when data is limited. Both of them utilized weakly-supervised training and validation data while the test data was curated through human supervision to assess the model performance. 
The first set consisted of a training set of size 95913, a validation set of size 31798 and a test set of size 1219. The second smaller dataset consisted of a training set of size 5000, a validation set of size 31798 and a test set of size 1219. It must be pointed out here that only the training data was changed between these two sets of experiments.

We intend to determine how well the Bidirectional LSTM (BiLSTM) networks with various static embeddings perform compared to the Bidirectional LSTM networks with the pretrained contextual embeddings from Elmo. For static embeddings, the 100-dimensional Glove embeddings were chosen as a tradeoff between expressiveness and availability of compute and memory resources. Along with the Glove embeddings, the 'Glove.twitter.100d' embeddings and 'Charngram.100d' embeddings were also evaluated. Training and validation accuracy on the weakly-supervised data is noted along with the final test accuracy on the fully-supervised test data. The history of training and validation accuracies on the smaller dataset are plotted in \autoref{glove_accuracy}, \autoref{glove_twitter_accuracy} and \autoref{charngrams_accuracy}. Since the size of the data was relatively small, the network had to be trained for 275 epochs to reach the performance obtained. Similarily, for the network with Elmo embeddings, training and validation accuracies were plotted in  \autoref{elmo_small_accuracy} for the smaller curated dataset and in  \autoref{elmo_large_accuracy} with the full training dataset. Although the results are similar to the static pretrained embeddings, Elmo reached this performance after 25 epochs for both datasets while the static embeddings had to be trained for 275 epochs on the smaller dataset to achieve similar performance. 

Statistics to illustrate the model performance of the static embeddings on the smaller dataset is plotted using the box plots shown in  \autoref{comparison_accuracy}, \autoref{comparison_f1}, \autoref{comparison_precision} and \autoref{comparison_recall} since this provides a better sense of the variance of the model performance than simply averaging out the results. For the network with Elmo embeddings, accuracy, recall, precision and F1-scores computed on the test set are calculated with both the smaller dataset and the larger dataset. Similar experiments were conducted on the network with static embeddings to assess how well it performs with limited data. As a result of the class imbalance in our training data, it is critical to evaluate performance using all of the metrics above. A summary of the averaged accuracy and F1-scores for all the models are presented in \autoref{tab:mean_acc_f1}. It is noteworthy to point out that in this regard, judging by the F1-score, the contextual embeddings provided by Elmo seems to perform marginally better than all of the static embeddings. It is also worth noting here that the performance of the network with Elmo embeddings is much more consistent when it is trained on the larger dataset. The threshold selected for class probabilities during classification while evaluating the above metrics was 0.5. However, ROC curves comparing the best-performing BiLSTM network with static embeddings (charngrams) and Elmo are plotted in \autoref{roc_curves} for comparison.

\begin{table}[]
\begin{tabular}{lll}
 \toprule
 & Accuracy &  F1-score  \\
 \toprule
 Glove(small data) & 82.165 &  80.422  \\
 Glove.twitter(small data) & 81.066 &  79.911  \\
 Charngrams(small data) & 80.098 &  79.232  \\
 Glove(full data) & 79.491 & 78.588 \\
 Glove.twitter(full data) & 79.688 & 78.713 \\
 Charngrams(full data) & 80.787 & 78.314 \\
 Elmo (small data) & 80.804 & 82.128 \\
 Elmo (large data) & 79.655 & 80.533 \\
 \toprule
\end{tabular}
\caption{\label{tab:mean_acc_f1}Mean metrics for all the models}
\end{table}

\subsection{Evaluation of visualization methods}

\begin{figure}[!h]
\includegraphics[width=\linewidth]{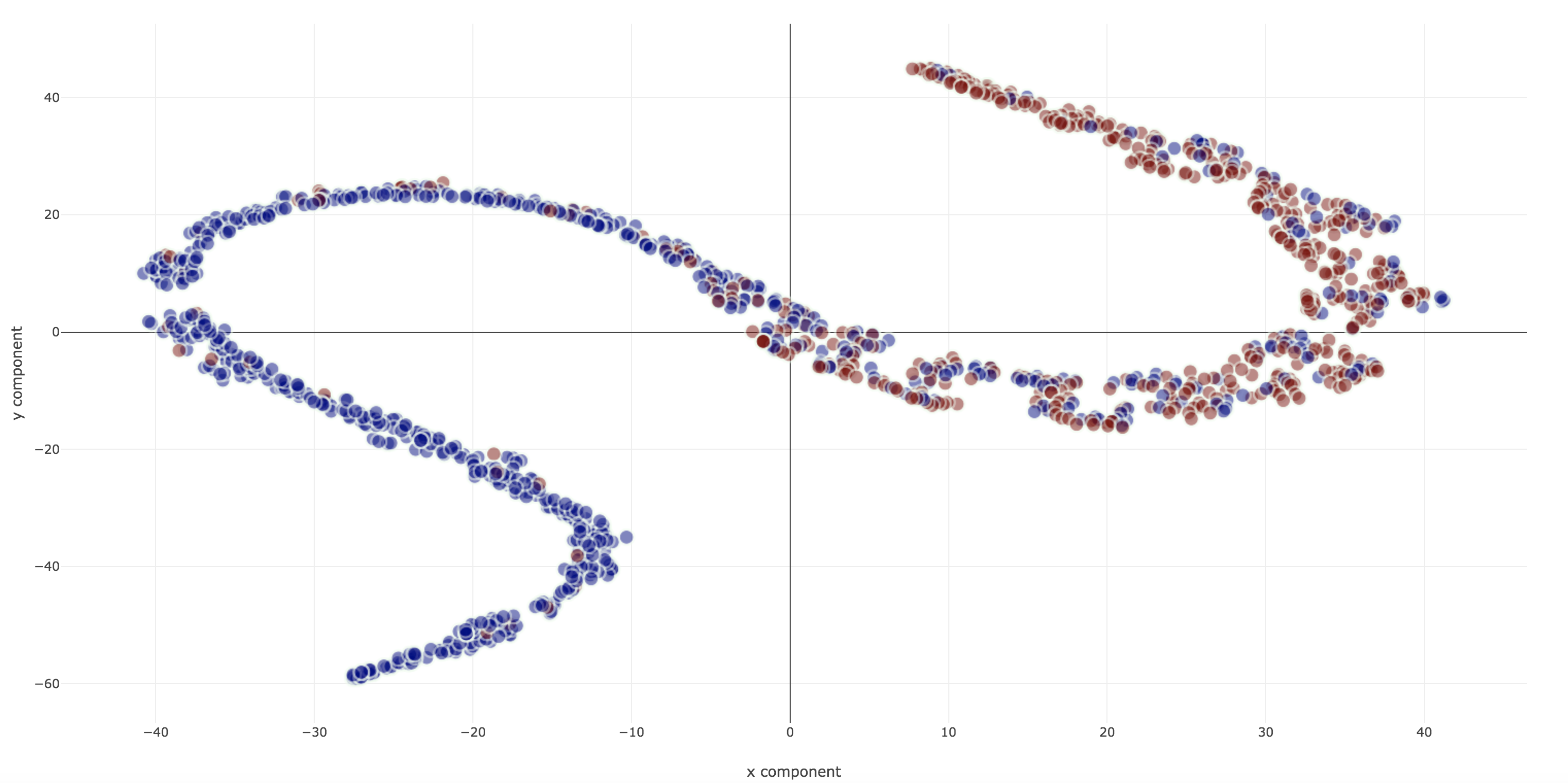}
\caption{Structure of t-SNE projections from the penultimate layer}
\label{tsne_proj}
\end{figure}

\begin{figure}[!h]
\begin{minipage}[t]{0.45\textwidth}
\includegraphics[width=\linewidth]{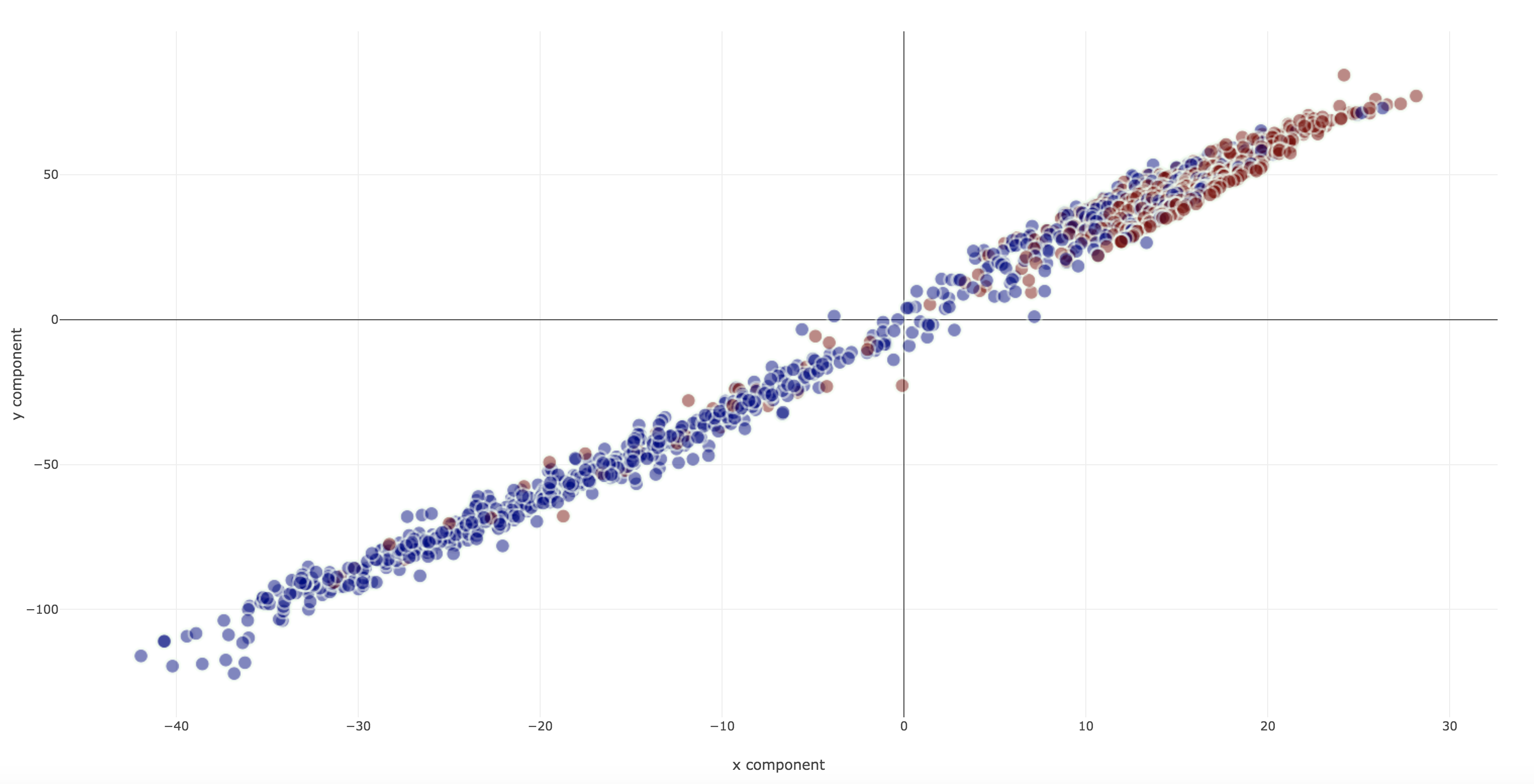}
\caption{Structure of MDS projections from the penultimate layer}
\label{mds_proj}
\end{minipage}
\end{figure}

\begin{figure}[!h]
\begin{minipage}[t]{0.45\textwidth}
\includegraphics[width=\linewidth]{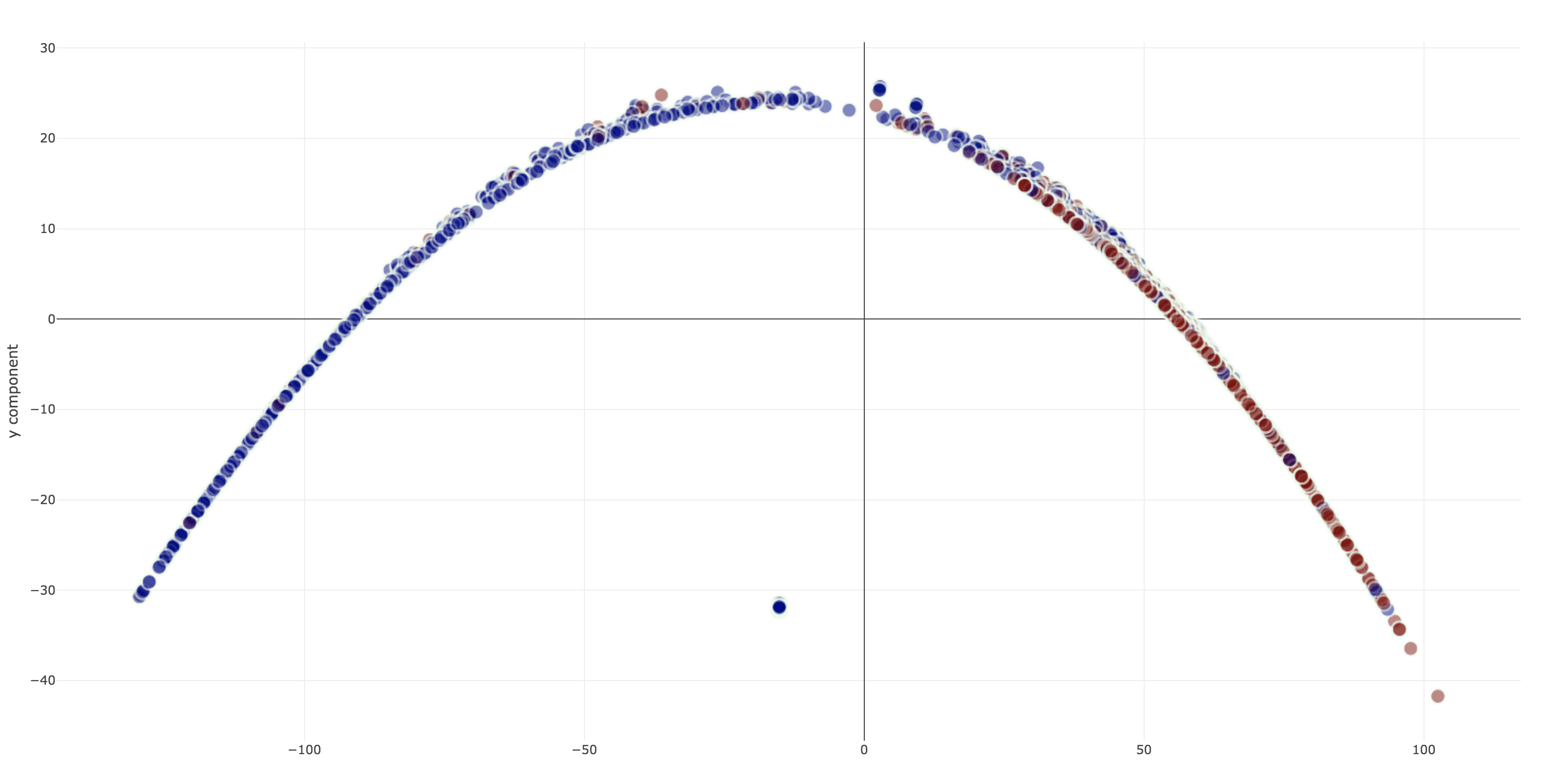}
\caption{Structure of Isomap projections from the penultimate layer}
\label{isomap_proj}
\end{minipage}
\end{figure}

\begin{figure}[!h]
\begin{minipage}[t]{0.45\textwidth}
\includegraphics[width=\linewidth]{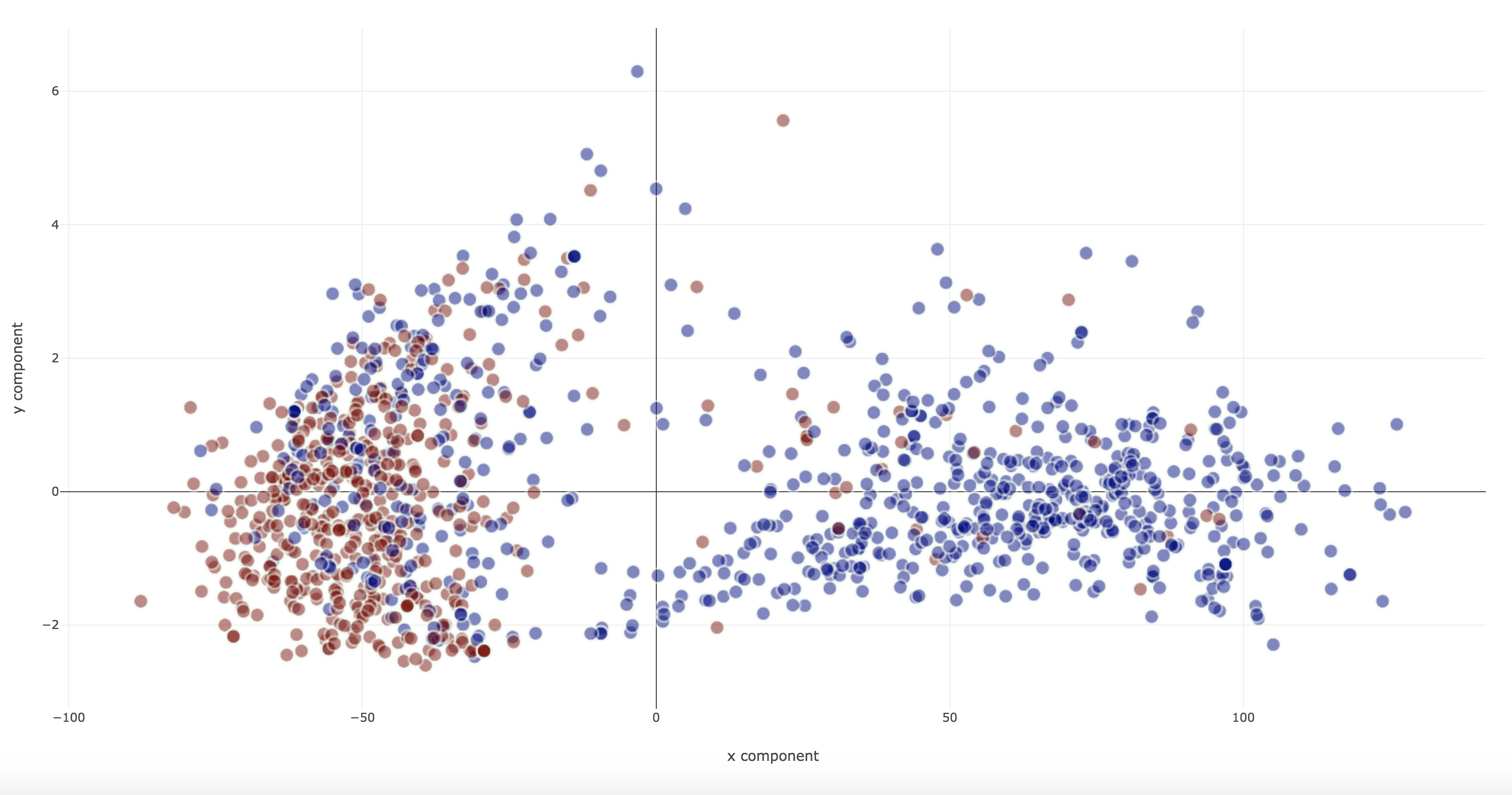}
\caption{Structure of PCA projections from the penultimate layer}
\label{pca_proj}
\end{minipage}
\end{figure}

Projections from the penultimate layer are dimension-reduced using PCA, MDS, Isomap and t-SNE to evaluate the suitability of these methods for representing the results of the networks and assessing political affiliation. t-SNE reveals clusters from high-dimensional manifolds while MDS groups entities based on similarities computed using a euclidean distance metric. The differences in the nature of the projections generated by these two techniques must be noted here; t-SNE tends to separate the entities (\autoref{tsne_proj}), reducing crowding, and forms distinct clusters while MDS projects the data along an axis (\autoref{mds_proj}). In a way, MDS allows quantification of stance as a function of `distance' along these axes. Even though t-SNE and MDS are used in the application, projections produced using both PCA (\autoref{pca_proj}) and Isomap (\autoref{isomap_proj}) were also considered. While clusters are visible in the PCA projection (\autoref{pca_proj}), it was much harder to identify the type `a' errors. Isomap, being a manifold projection technique, produces more distinct aggregation of entities and projects them along a non-linear axis. In this regard, Isomap (\autoref{isomap_proj}) can be seen as a better technique for capturing non-linear relationships in the high-dimensional data while still providing a way to quantify affiliations in the form of geodesic distances along the axes formed by the projection.

For the purpose of determining political affiliation, it was determined that the projections produced by MDS were more cognitively efficient, or had better visualization efficiency. 
 \subsection{User evaluation}

\begin{figure}[!h]
\includegraphics[width=\linewidth]{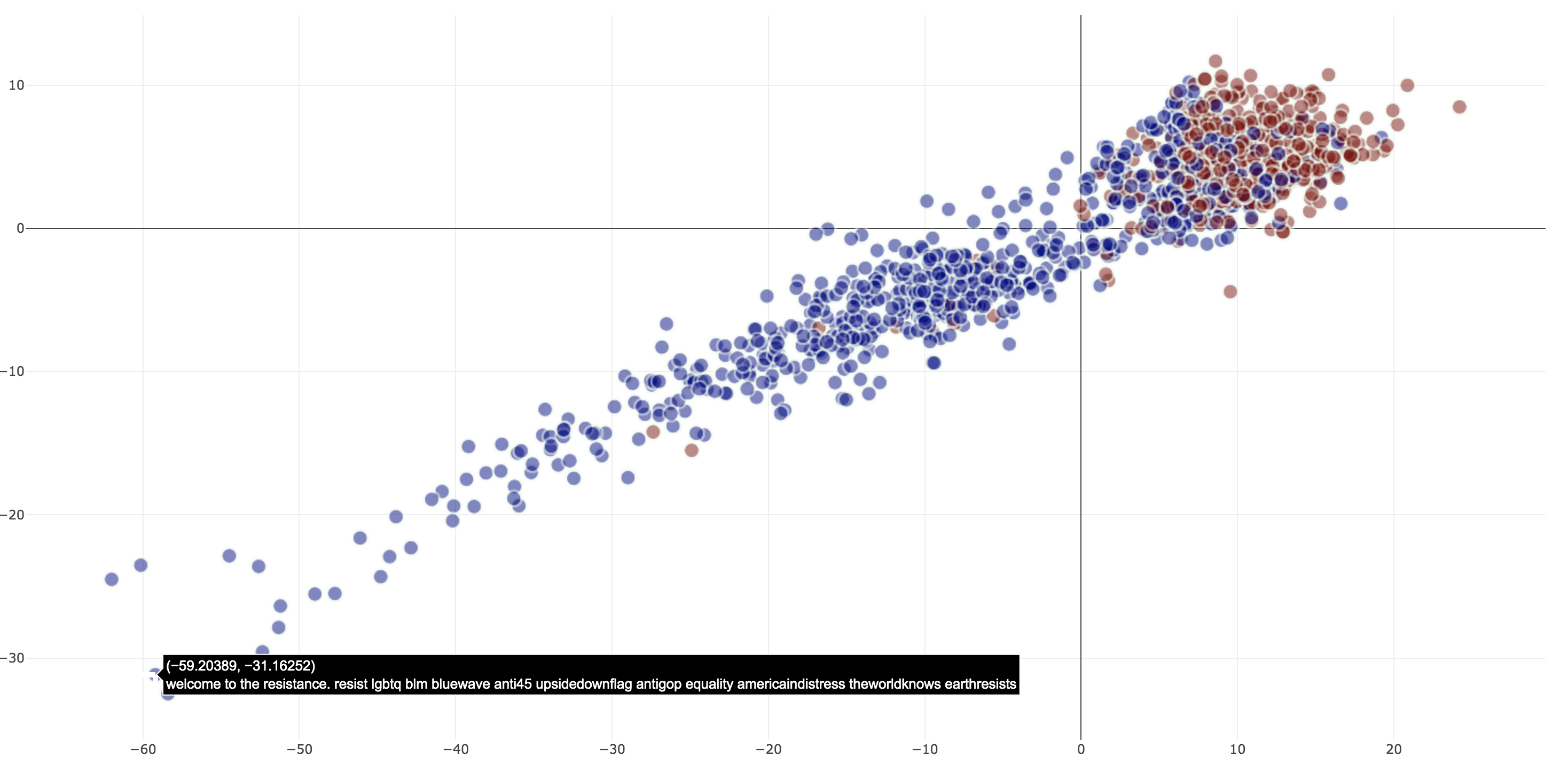}
\caption{Liberal ideology positioned at the bottom left}
\label{mds_liberal}
\end{figure}

\begin{figure}[!h]
\begin{minipage}[t]{0.45\textwidth}
\includegraphics[width=\linewidth]{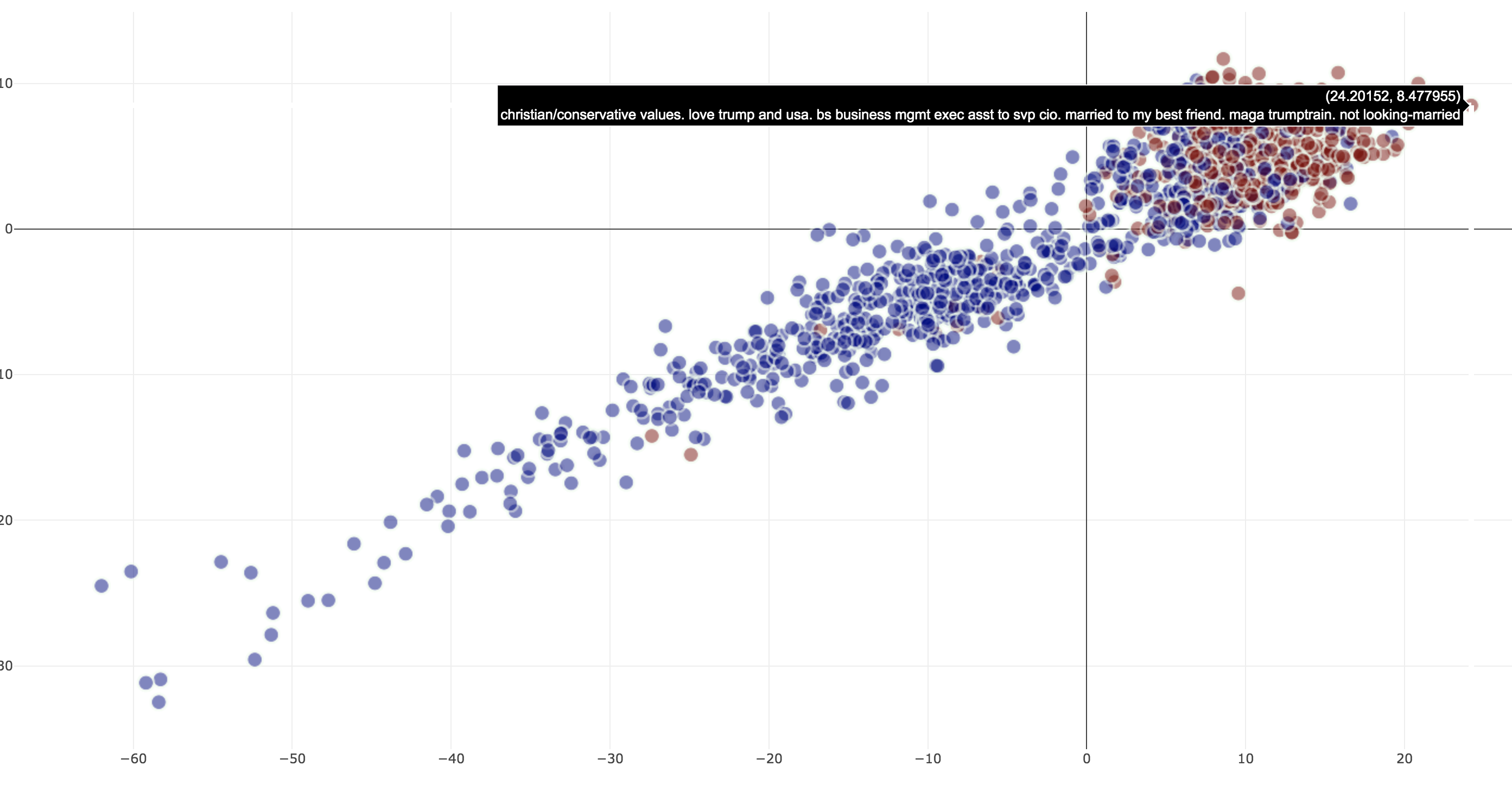}
\caption{Conservative ideology positioned at top right}
\label{mds_conservative}
\end{minipage}
\end{figure}

\begin{figure}[!h]
\begin{minipage}[t]{0.45\textwidth}
\includegraphics[width=\linewidth]{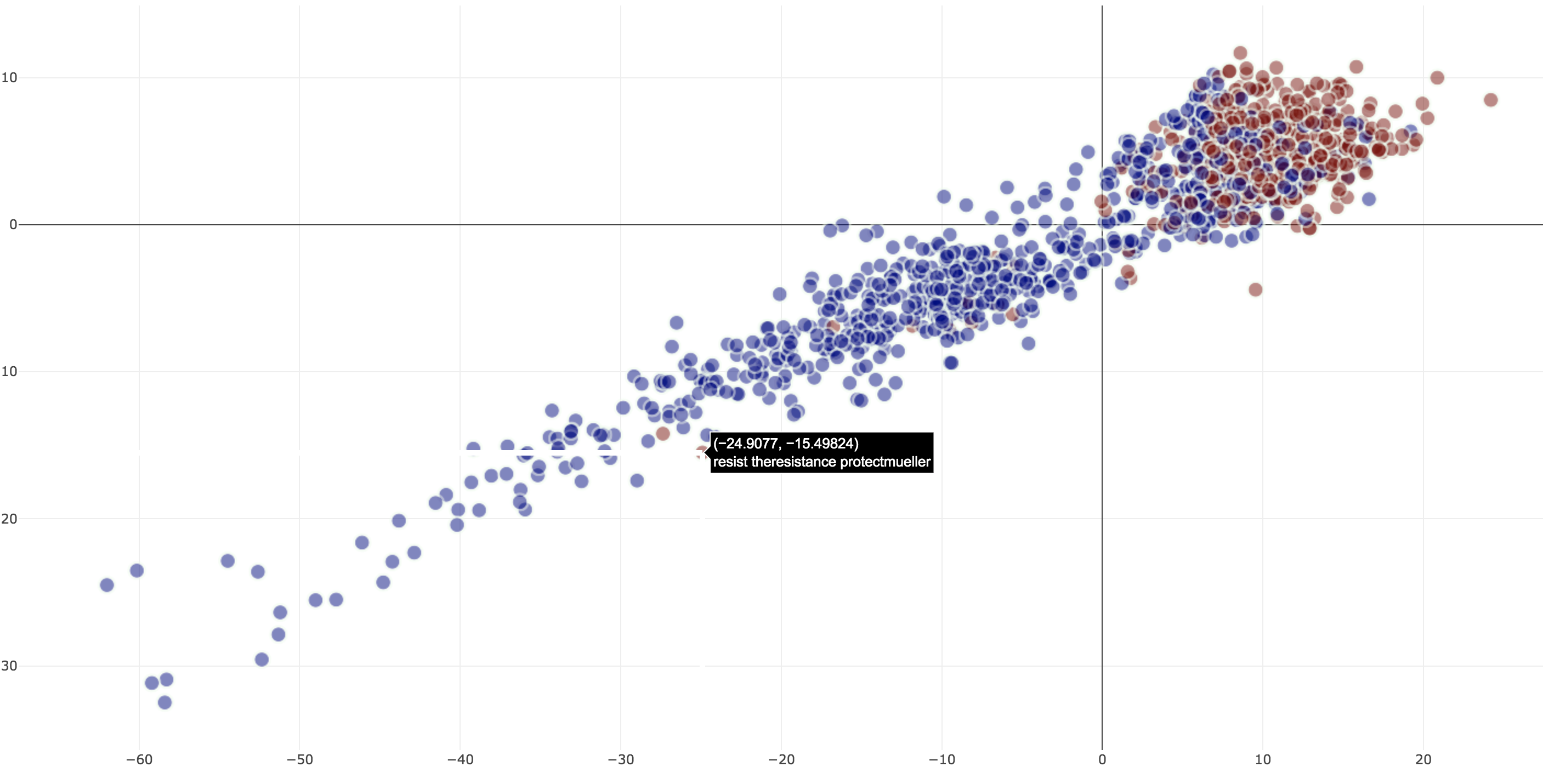}
\caption{Identification of type `a' error with MDS}
\label{mds_error}
\end{minipage}
\end{figure}

For the purpose of determining political affiliation, it was determined that the projections produced by MDS were more cognitively efficient.  \autoref{mds_liberal} and  \autoref{mds_conservative} indicate how affiliation is projected along the orientation of the cluster with liberal ideology being represented at the bottom left and conservative ideology being represented at the top right. It is interesting to note in \autoref{mds_error} how this projection was able to identify a type 'a' error of an incorrect label in the corpus. Even though the same error existed in the PCA, t-SNE and Isomap projections, it was easier to identify them in the plot generated by MDS. It must be noted here that the identification of type `a' errors was done on the annotated test data by creating a perturbed test set. This perturbed test set was essentially the same test set that had 5 record labels changed. This was done to assess if the projections enabled the user to clean the perturbed data more efficiently, compared to simply inspecting the entire test set, by helping to identify the errors. 

From the perspective of using contextual embeddings eith Elmo no significant difference was observed in the structure  of the visualizations compared to the static embeddings. The results from the Attention weights for each document informed the user how much the network emphasized each token in a document. The hashtags, as expected, displayed higher weights however other words were less expected as was shown in \autoref{attention}. In a way, this provided a token-level feature extraction identifying key attributes in a category through weak supervision. Additionally, it also emphasized the need for a more rigorous document tokenization/cleaning to avoid weighting spurious tokens.

Preliminary evaluations with a domain expert suggest improved efficiency for corpus label cleaning in single pass trials. For MDS, which proved to be the most efficient form of projection for both the static and contextual embeddings, the user spent a combined average time of 2.2 minutes for both the review of the visualizations and the decision-making process to identify 5 errors. Isomap was close behind in identifying 5 errors but it took on average 2.4 minutes, followed by t-SNE which showed promising clustering properties. PCA performed the worst at being able to identify the 5 inaccurate labels.

\section{Conclusion and Future Work}
The feasibility of labelling documents using weakly-supervised data was evaluated. Minimal human supervision was required due to the ability of the trained DNN to produce projections that helped the human expert to quickly identify errors. This is in contrast with the laborious process of inspecting every document in the corpus. The current configuration has helped to improve the labeling accuracy, but more tests need to be performed with the use of additional pretrained contextual embedding layers. Various network configurations using automated hyperparameter and architecture identification techniques such as AutoML are also being evaluated to improve performance. Embedding token class information into the documents, as opposed to using the raw documents for training, might also be beneficial for the DNN. 

The application can also be extended to work in a distributed fashion so that users at various sites can collaborate on decision-making with regards to a corpus of documents. A related non-trivial issue is the matter of document cleaning; iterative passes through the pipeline informs the user of better strategies to employ during document pre-processing. Finally, scalability to larger datasets has not been addressed in this work, which the authors hope to assess soon.

\bibliographystyle{abbrv-doi}
%\bibliographystyle{abbrv-doi-narrow}
%\bibliographystyle{abbrv-doi-hyperref}
%\bibliographystyle{abbrv-doi-hyperref-narrow}

%\bibliography{template}

\end{document}